%% file: main.tex
\documentclass[10pt,twocolumn,letterpaper]{article}

\usepackage{iccv}              
\usepackage[hang,flushmargin]{footmisc}
\newcommand\blfootnote[1]{%
    \begingroup 
    \renewcommand\thefootnote{} \footnote{#1}%
    \addtocounter{footnote}{-1}%
    \endgroup 
}


\definecolor{iccvblue}{rgb}{0.21,0.49,0.74}
\usepackage[pagebackref,breaklinks,colorlinks,allcolors=iccvblue]{hyperref}

\def\paperID{4446} 
\def\confName{ICCV}
\def\confYear{2025}

\title{RAGDiffusion: Faithful Cloth Generation via External Knowledge Assimilation}

\author{ 
Xianfeng Tan$^{2*}$, Yuhan Li$^{1*\dag}$, Wenxiang Shang$^{2}$, Yubo Wu$^{2}$, Jian Wang$^{2}$, \\
Xuanhong Chen$^{1}$, Yi Zhang$^{1}$, Hangcheng Zhu$^{2\ddag}$, Bingbing Ni$^{1\ddag}$
\\[2pt] 
$^{1}$Shanghai Jiao Tong University  \hspace{2em}  $^{2}$Alibaba Group  
\\[2pt] 
\small
\url{https://colorful-liyu.github.io/RAGDiffusion-page/} \\
}

\begin{document}
\maketitle

\blfootnote{* Joint first authors. \\
\dag  Work done during an internship at Alibaba. \\
 \ddag  Joint corresponding authors. \\ }

\input{sec/0_abstract}    
\input{sec/1_intro}

\input{sec/2_relatedwork}

\input{sec/3_method}  
\input{sec/4_expriment}

\input{sec/5_conclusion}

\input{sec/X_supp}  
{
    \small
    \bibliographystyle{ieeenat_fullname}
    \bibliography{main}
}

\end{document}

%% file: sec/0_abstract.tex
 \vspace{-5mm}
\begin{abstract}
Standard clothing asset generation involves restoring forward-facing flat-lay garment images displayed on a clear background by extracting clothing information from diverse real-world contexts, which presents significant challenges due to highly standardized structure sampling distributions and clothing semantic absence in complex scenarios. Existing models have limited spatial perception, often exhibiting structural hallucinations and texture distortion in this high-specification generative task. To address this issue, we propose a novel Retrieval-Augmented Generation (RAG) framework, termed RAGDiffusion, to enhance structure determinacy and mitigate hallucinations by assimilating knowledge from language models and external databases. RAGDiffusion consists of two processes: (1) Retrieval-based structure aggregation, which employs contrastive learning and a Structure Locally Linear Embedding (SLLE) to derive global structure and spatial landmarks, providing both soft and hard guidance to counteract structural ambiguities; and (2) Omni-level faithful garment generation, which introduces a coarse-to-fine texture alignment that ensures fidelity in pattern and detail components within the diffusing. Extensive experiments on challenging real-world datasets demonstrate that RAGDiffusion synthesizes structurally and texture-faithful clothing assets with significant performance improvements, representing a pioneering effort in high-specification faithful generation with RAG to confront intrinsic hallucinations and enhance fidelity.
\end{abstract}

%% file: sec/1_intro.tex
 \vspace{-5mm}
\section{Introduction}
 \vspace{-2mm}

\label{sec:intro}

\begin{figure}[t]
  \centering
  \includegraphics[width=1\linewidth]{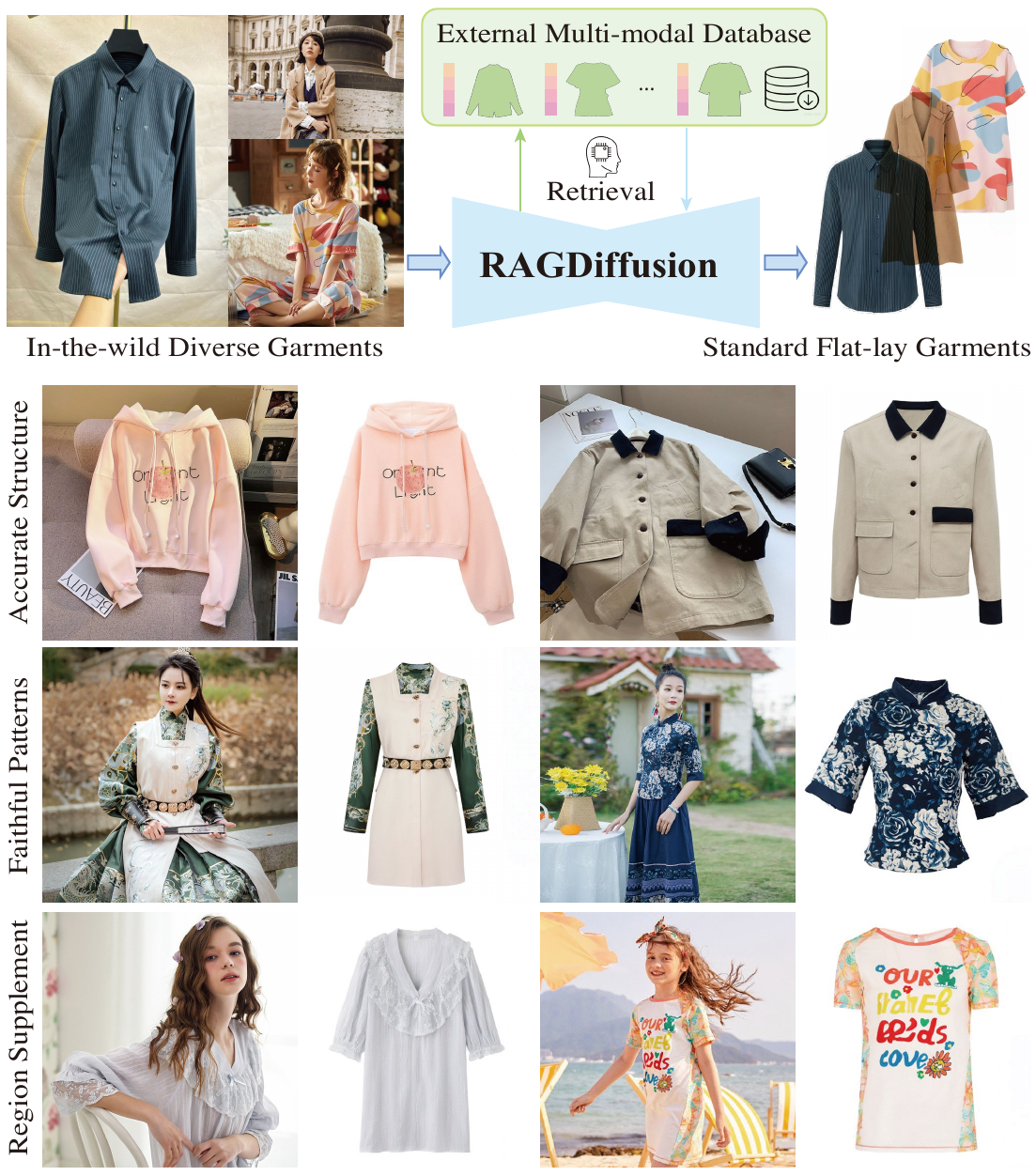}
    \vspace{-5mm}
  \caption{RAGDiffusion synthesizes structurally and pattern-wise faithful standard garments by assimilating retrieved knowledge. }
  \vspace{-6mm}
  \label{fig: motivation}
\end{figure}

While the high quality of images generated by diffusion models~\cite{ho2020ddpm, song2020ddim} has significantly lowered the barriers for individuals to engage in content creation, the generation of 2D standard clothing assets from in-the-wild data has been relatively underexplored~\cite{shen2024igr}. Standard clothing asset generation refers to generating forward-facing flat-lay garment images~\cite{velioglu2024tryoffdiff} on a clear background by recovering clothing information from arbitrary real-world scenes (\eg, garments hung on hangers, worn by models, or casually laid on chairs in \cref{{fig: motivation}}). Standard garments serve as a crucial intermediary variable, connecting various downstream applications such as e-commerce catalogs maintenance, garment design, product marketing and virtual fitting~\cite{he2024dresscode, zhang2023diffcloth, zhang2022armani}.

The garment generation task presents dual technical challenges distinct from conventional virtual try-on paradigms. First, standard clothing generation necessitates \emph{standard structure and frontal positioning} while inherently \textbf{limiting stylistic flexibility}, constrained by the narrow manifold of \textbf{high-specification garment distributions}~\cite{zhang2024garmentaligner}. Second, \textbf{the complexity of in-the-wild conditions} involving occlusions, multi-layer outfits, truncations, or extreme lateral viewpoints notably exacerbates \emph{structural uncertainty and pattern ambiguity}. Collectively, experimental evaluations reveal that existing methods~\cite{xarchakos2024tryoffanyone, hu2023animate} fail to achieve practical usability thresholds with structural and texture distortion, particularly manifesting a pathological \emph{\textbf{structure hallucinations}} in \emph{inaccurate structure, fitness and length estimation} under challenging scenes (see ~\cref{fig: main} and ~\cref{fig: ablation}), which we attribute to the limited spatial reasoning capacity inherent in Stable Diffusion (SD) architectures.

Despite its critical commercial relevance, the garment restoration problem remains insufficiently addressed in literature~\cite{zeng2020tilegan, velioglu2024tryoffdiff, xarchakos2024tryoffanyone, shen2024igr}. Early GAN-based frameworks~\cite{zeng2020tilegan} suffer from pattern distortions constrained by their generative priors. Recent approaches~\cite{velioglu2024tryoffdiff, xarchakos2024tryoffanyone, shen2024igr} employing pre-trained SD exhibit several limitations: (i) inadequate handling of structural hallucinations in hard cases (occluded, multi-layer, \etc), (ii) commercially unfaithful logo/texture reproduction which is extremely sensitive to e-commerce sellers. In summary, no existing solution achieves the dual objectives of high-structural integrity and photorealistic texture synthesis under real-world complexity.

We propose RAGDiffusion, as a Retrieval-Augmented Generation (RAG)~\cite{ram2023conrag2} paradigm including two processes: information aggregation based on retrieval, and conditional generation with omni-level fidelity, to address the issue of structural and texture distortion respectively.
(1) {\textbf{Retrieval-based Structure Aggregation}: Our key insight lies in enhancing \emph{\textbf{structural determinacy}} through the assimilating of external structural landmarks and world knowledge, thereby rectifying visual models' misconceptions regarding garment structure, fitness and proportions.  During RAG process, contrastive learning~\cite{he2020moco, chen2020simclr} is firstly introduced to train a dual tower network to extract multi-modal structure embeddings from images of two branches as well as attributes derived from a frozen large language model (LLM)~\cite{bai2023qwen, achiam2023gpt}. Considering potential encoding bias in real-world data, we propose a structure retrieval named Structure Locally Linear Embedding (SLLE) to project the predicted structure embedding towards a standard manifold~\cite{roweis2000lle, ye2023geneface++} as well as offer a silhouette landmark. The remapped latent structure embedding and the landmarks provide \emph{soft} and \emph{hard} structure guidance respectively through Embedding Prompt Adapter and Landmark Guider to eliminate structural hallucinations.
(2) {\textbf{Omni-level Faithful Garment Generation}: RAGDiffusion establishes coarse-to-fine texture alignment for \emph{\textbf{pattern faithfulness}} and \emph{\textbf{detail faithfulness}} for generated garment. Specifically, we ensure the generated pattern matches the conditioning through ReferenceNet~\cite{hu2023animate}, a conditioning UNet isomorphic to the denoising UNet. However, problematic detail and logo artifacts persist partially due to resolution limitations of VAE decoders, rendering the generated images commercially unusable for e-commerce sellers, which is an under-addressed issue across prior research~\cite{velioglu2024tryoffdiff, xarchakos2024tryoffanyone, shen2024igr}. To mitigate reconstruction distortions inherent in original VAE~\cite{kingma2013vae}, we propose Parameter Gradual Encoding Adaptation (PGEA) to adapt the SDXL~\cite{podell2023sdxl} backbone to a more powerful VAE.

To the best of our knowledge, RAGDiffusion stands as a pioneering multi-modal RAG method to solve intrinsic hallucination and unfaithfulness during image synthesis. 
Furthermore, we discover two additional benefits brought by RAG: \textbf{zero-shot generalizability} for unseen scenarios via retrieval database expansion, and \textbf{human-interpretable control} mechanisms through landmark manipulation. The core contributions of this work are threefold:

\begin{itemize}
\item We propose RAGDiffusion, a RAG framework for clothing generation with representation learning and SLLE to offer structure determinacy and eliminate hallucination.
\item We employ a coarse-to-fine texture alignment in RAGDiffusion, addressing pattern and detail aspects, ensuring faithfulness of clothing on this high-specification task.
\item Comprehensive experiments on challenging in-the-wild sets demonstrate RAGDiffusion is capable of synthesizing both structure and texture faithful clothing assets, outperforming current methods by a substantial margin. Furthermore, the ablation study has validated the effectiveness of different parts of RAGDiffusion.
\end{itemize}

%% file: sec/2_relatedwork.tex
\section{Related works}
 \vspace{-2mm}

\noindent \textbf{Retrieval-augmented generative models.} 
Retrieval-augmented strategies leverage external databases to enhance the capabilities of generative models across a variety of tasks. For instance, the RETRO~\cite{RETRO} modifies an LLM to effectively utilize external databases. In the realm of image synthesis, retrieval has been employed in both GANs~\cite{tseng2020retrievegan, casanova2021instance} and diffusion models~\cite{blattmann2022retrievaldiff, sheynin2022knndiffusion}, accommodating 3D generation~\cite{seo2024retrieval3d}, video generation~\cite{zhang2023remodiffuse} and artistic styles~\cite{rombach2022retrievalArtistic}. The essence of these works lies in retrieving similar images to serve as mimetic references for generating specified content, particularly in cases where models are insufficiently trained. Our RAGDiffusion not only builds upon this generalizability through convenient retrieval expansion, but also aggregates structural information, thereby mitigating intrinsic hallucinations on the high-standard tasks.

\begin{figure*}[t]
  \centering
  \includegraphics[width=0.95\linewidth]{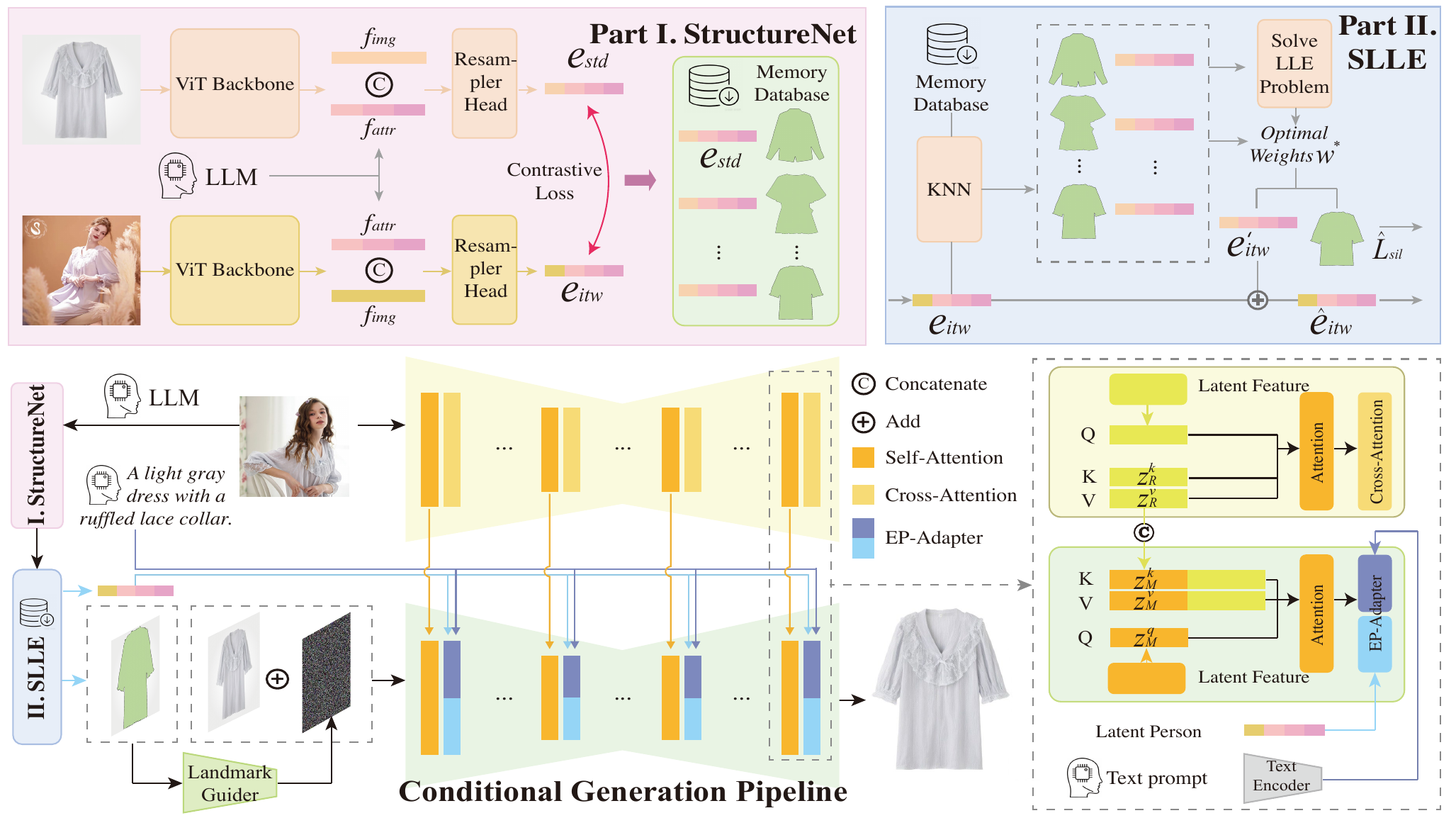}
  \vspace{-2mm}
  \caption{Overall framework of RAGDiffusion. StructureNet provides latent structural embeddings, while SLLE facilitates the embedding fusion along with landmark retrieval. The generative model synthesizes multiple conditions to achieve omni-level high-fidelity generation.  }
  \vspace{-4mm}
  \label{fig: framework}
\end{figure*}

\noindent \textbf{Controllable text-to-image diffusion models.} 
To attain conditional control in text-to-image diffusion models, ControlNet~\cite{zhang2023controlnet}, T2I-Adapter~\cite{mou2024t2iadapter} and IP-Adapter~\cite{ye2023ipadapter} incorporate additional trainable modules to fuse conditions on feature maps. Additionally, recent investigations have utilized a variety of prompt engineering techniques~\cite{li2023gligen, yang2023reco, zhang2023controllablegpt4} and implemented cross-attention constraints~\cite{chen2024layoutcontrol, xie2023boxdiff, ju2023humansd, zhang2024boow} to facilitate more controllable generative processes. Furthermore, some studies further investigate multi-condition or multi-modal generation~\cite{hu2023cocktail, qin2023unicontrol, zhao2024uni, li2024anyfit}. However, these methods rely on the associative capabilities of stable diffusion, which cannot guarantee fidelity, particularly with the highly standardized sampling of standard garments.

\noindent \textbf{Garment restoration.}  
Standard clothing generation involves restoring flat-lay garment images on from real-world contexts. TileGAN~\cite{zeng2020tilegan} pioneered a two-stage GAN~\cite{goodfellow2014GAN} framework, suffering from backbone capability limitations. Recent works including TryOffDiff~\cite{velioglu2024tryoffdiff}, TryOffAnyone~\cite{xarchakos2024tryoffanyone}, and IGR~\cite{shen2024igr} have adopted pretrained SD as base backbones, yielding obvious improvements through enhanced generative priors. Specifically, TryOffDiff utilizes SigLIP~\cite{zhai2023sigmoid} to extract garment features; TryOffAnyone employs direct image concatenation for conditional generation; IGR implements ReferenceNet-inspired~\cite{hu2023animate} architectures for pattern consistency. However, these methods exhibit three critical shortcomings: (1) They ignore structural hallucinations and distortions arising from real-world complexities (multi-layer, extreme viewpoints, \etc). (2) Their detail fidelity fails to meet commercial standards. (3) Adaptation to new garment categories/scenarios requires costly paired data and retraining. Our RAGDiffusion systematically resolves these limitations through RAG paradigm and a coarse-to-fine alignment pipeline.

%% file: sec/3_method.tex
\section{Method}
 \vspace{-2mm}

An overview of the RAGDiffusion is presented in~\cref{fig: framework}. The backbone of RAGDiffusion employs the SDXL~\cite{rombach2022ldm}, with the preliminary detailed in Appendix. Given an in-the-wild clothing image $x_{itw} \in \mathbb{R}^{H\times W\times 3}$, RAGDiffusion is aimed to generate an authentic standard flat lay in-shop garment image $x_{std}$. A dual tower StructureNet extracts latent structure embeddings by contrastive learning as detailed in \cref{sec: emb}, and SLLE retrieves and fuses structure embeddings and landmarks as in \cref{sec: SLLE}. Coarse-to-fine alignment generation involves ReferenceNet,and PGEA for pattern, and decoding faithfulness described in ~\cref{sec: omni}.

\subsection{Dual-tower embeddings extraction}
\label{sec: emb}
 \vspace{-2mm}

To eliminate structural hallucinations, a straightforward approach is to extract structural features from images and feed them into generative networks, as seen in StyleGAN~\cite{karras2019style} and LADI-VTON~\cite{morelli2023ladi}, which use additional latent coding to module conditions. Accordingly, we extract latent structure embeddings through contrastive learning~\cite{radford2021clip}, using garments that share similar canny but are texture-dissimilar as training pairs. This process exploits the structural similarities between specialized pairs, which are not emphasized during the conventional training of diffusion.

\noindent \textbf{Contrastive learning.} Given a batch of $N$ (in-the-wild clothing $x_{itw}$, standard clothing $x_{std}$) pairs, StructureNet $g_{\theta}$ is trained to predict which of the $N \times N$ possible ($x_{itw}$, $x_{std}$) pairings across a batch actually occurred according to structure similarity. To do this, StructureNet learns a multi-modal embedding space by jointly training a dual tower encoder tailored for $x_{itw}$ and $x_{std}$ to maximize the cosine similarity of the embeddings of the $N$ real pairs (marked with superscript $+$) in the batch while minimizing the similarity of the $N^2-N$ incorrect pairs(marked with superscript $-$). To be specific, the LLM is used to extract 10 types of discrete attributes of clothing (\eg, Category, FitType, CollarTechnique, IfTopTuckIn), which incorporates general in-context image knowledge from the language model~\cite{achiam2023gpt, bai2023qwen}, \emph{eliminating semantic deficiencies and ambiguities present in vision}. Then we assign a learnable embedding $f_{attr}$ to each discrete attribute. The image structure features $f_{img}$ are extracted by the twin ViT~\cite{dosovitskiy2020vit} encoder and concatenated with the attributes embeddings $f_{attr}$ to form the final latent structure embeddings $e$ after non-linear Resampler~\cite{ye2023ipadapter} head. We optimize the InforNCE loss~\cite{chen2020simclr} over these latent structure embeddings $\left \{ e_{itw}, e_{std} \right \} $ from $\left \{ x_{itw}, x_{std} \right \}$ as:
\begin{equation}
\small
\mathcal{L} = \frac{-1}{N}\sum_{i=1}^{N}\text{log}\frac{\text{exp}(e_{itw, i}\odot e_{std}^+/\tau)} {\text{exp}(e_{itw, i}\odot e_{std}^+/\tau)+ {\textstyle \sum_{e_{std}^-}} \text{exp}(e_{itw, i}\odot e_{std}^-/\tau)},
\end{equation}
where $\odot$ is cosine similarity between two vectors, $N$ is the batch size, and $\tau$ is a temperature scalar.

\subsection{Retrieval-augmented SLLE}
\label{sec: SLLE}
 \vspace{-2mm}

By integrating structural knowledge from StructureNet into the SD, we have improved the style and structure of generated flat-lay clothing as shown in ~\cref{sec: ablation}. However, due to the limited spatial perception inherent in the SD~\cite{borji2023qualitative, liu2023intriguing}, it often struggles to accurately represent the length and contours of clothing, particularly in hard cases (occlusions, multi-layer, lateral viewpoints, \etc), as in \cref{fig: ablation}.
Furthermore, since StructureNet is trained on a limited set of contrastive pairs, it may perform poorly with out-of-distribution samples, making it unreliable for real-world applications.

We have observed that creating a comprehensive database of standard flat-lay clothing for retrieval is easier than gathering comprehensive in-the-wild samples (numerous scenarios). In this context, we set standard clothing $e_{std}$ as basic vectors and propose Structure Locally Linear Embedding (SLLE), which is a manifold projection ~\cite{roweis2000lle} to correct each predicted structure embedding $e_{itw}$ into the target space of standard embedding $e_{std}$ to mitigate potential error. Meanwhile, the retrieval database provides structure landmarks to strengthen explicit spatial constraints.

\noindent \textbf{Memory database.} 
To execute SLLE, we establish a retrieval memory database. We first encode the collected standard flat-lay clothing image dataset into a series of latent structure embeddings $e_{std}$ with StructureNet $g_{\theta}$. The silhouette landmarks $L_{sil}$ are also extracted to designate areas for generated content. Thus an external memory database $\mathcal{D}$ that consists of embedding-landmark pairs ($e_{std}$, $L_{sil}$) is obtained. These structure embeddings are utilized in matching algorithms~\cite{guo2003knn}, enabling a given in-the-wild garment to find the most compatible standard features $e_{std}$ and landmarks during inference.  Notice that landmarks can be any structural figure. Here, we use the outline of clothing, primarily to tackle the challenges of limited spatial perception in SD (\eg length and contour)~\cite{borji2023qualitative, liu2023intriguing}.

\noindent \textbf{Structure LLE algorithm.} 
In this process, retrieval-augmented SLLE drags in-the-wild embedding closer to the standard flat-lay garment embeddings to void outliers during inference, as well as offer silhouette landmarks. Motivated by the successful practice of classic locally linear embedding in~\cite{blanz2003face}, we assume that the garment embedding or landmark and its nearby points are locally linear on the manifold, eliminating the need to project them into a higher-dimensional space as vanilla LLE~\cite{roweis2000lle} does. Specifically, given an extracted garment embedding $e_{itw}$, the goal of SLLE is to reconstruct embedding $e_{itw}'$ with standard embeddings $e_{std}$ as basis vectors. We start by searching the \( K \) nearest standard embeddings \( \{ e_{std}^1, ...,e_{std}^K \} \) from the standard garment memory database \( \mathcal{D} \) using cosine similarity. Thus $K$ corresponding flat-lay cloth silhouette landmarks $\left \{ L_{sil}^1, ...,L_{sil}^K \right \}$ are obtained as well. Next, a linear combination of these neighbors is sought to reconstruct $e_{itw}'$ by minimizing the error $\left \| e_{itw}'-e_{itw} \right \| _2$, which could be formulated as the least-squared optimization problem:
\begin{equation}
\min \left \| e_{itw}-{\textstyle \sum_{i=1}^{K}w_i \cdot e_{std}^i} \right \| _2, \quad s.t.\textstyle \sum_{i=1}^{K}w_i=1,
\label{eq: least_squared}
\end{equation}
where $w_i$ is the barycentric weight of the $i$-th nearest embedding $e_{std}^i$. The optimal weights \( \{ w_{1}, \ldots, w_{K} \} \) can be determined by solving \cref{eq: least_squared}. Subsequently, we can reconstruct the shape embedding as \( e_{itw}' = \sum_{i=1}^{K} w_i^* \cdot e_{std}^i \). Ideally, the reconstructed embedding \( e_{itw}' \) serves as an in-domain data point that retains structural accuracy and appropriate fitness, albeit with some information loss. In practice, we use a linear combination of original structure embedding $e_{itw}$ and reconstructed one $e_{itw}'$ as the final structure representation during inference:
\begin{equation}
\small
\hat{e}_{itw} = \alpha \cdot {\textstyle \sum_{i=1}^{K}w_i^* \cdot e_{std}^i}+ (1-\alpha )\cdot e_{itw},
\label{eq: fuse}
\end{equation}
where $\alpha \in [0, 1]$ controls the trade-off and is set to be $0.5$. The final landmark $\hat{L_{sil}}$ is also fused with optimal weights.

\noindent \textbf{Structure conditioning.} 
Inspired by IP-Adapter~\cite{ye2023ipadapter}, we adopt a similar \emph{\textbf{Embedding Prompt Adapter (EP-Adapter)}} to condition the high-level structure semantics. While maintaining the text branch unchanged, fused structure embeddings $\hat{e}_{itw}$ after SLLE are fed into additional projection layers to generate key and value matrices for the structure representations. Two parallel cross-attention layers process the \emph{text modality} and the \emph{structure embedding modality}, with results being summed to replace the original single text one: $\mathsf{Attention}(Q, K_{text}, V_{text})+\mathsf{Attention}(Q, K_{emb}, V_{emb})$. The EP-Adapter enhances the global/inner structural faithfulness by incorporating prior knowledge from LLM and structural training pairs as a soft constraint. Meanwhile, the contour landmark $\hat{L_{sil}}$ involves external spatial structural information. A Landmark Guider~\cite{hu2023animate} with $4$ convolution layers ($4\times 4$ kernels, $2\times 2$ strides,  $16,32,64,128$ channels) is incorporated to align the silhouette mask with $z_t$ as an explicit and hard constrain.

\subsection{Omni-Level faithful garment generation}
\label{sec: omni}
 \vspace{-1mm}
 
\noindent \textbf{Pattern faithfulness.}
Inspired by success in human editing~\cite{hu2023animate, chang2023magicdance}, we introduce an additional UNet encoder (\ie \emph{\textbf{ReferenceNet}}~\cite{hu2023animate}) to precisely preserve the fine-grained details of clothing assets, which is isomorphic to the main generative UNet (\ie MainNet) and shares same initial parameter weights. The latent of in-the-wild garment image is passed through ReferenceNet to obtain the intermediate \emph{key} and \emph{value} features $\left \{ z_{R}^k, z_{R}^v \right \} \in \mathbb{R}^{N\times l\times d}$ in self-attention, which are concatenated with the features $\left \{ z_{M}^k, z_{M}^v \right \} \in \mathbb{R}^{N\times l\times d}$ from the MainNet along the $l$ dimension to obtain the final $\left \{ z_{C}^{k}, z_{C}^{v} \right \} \in \mathbb{R}^{N\times 2 l\times d}$. Then we compute the self-attention on the concatenated features as:
\begin{equation}
\text{Attention}\left \{ z_{M}^q, z_{C}^k, z_{C}^v \right \} = \text{softmax}(\frac{z_{M}^q z_{C}^{kT}}{\sqrt{d}})z_{C}^v,
\end{equation}
where $z_{M}^q$ represents \emph{Query} features in self-attention from MainNet. ReferenceNet plays an important role in texture preserving of garments when it has complicated patterns. 


\noindent \textbf{Detail faithfulness.}
The cloth restoration technique primarily serves e-commerce sellers for product catalog maintenance and marketing purposes. The requirement for brand logo fidelity is so stringent that existing methods' results~\cite{velioglu2024tryoffdiff, xarchakos2024tryoffanyone, shen2024igr} are practically unusable even with ReferenceNet, partially because of the fine-grained logo decoding degradation caused by the high compression ratio of SDXL's VAE, as shown in~\cref{fig: ablation}. Although the VAE in FLUX~\cite{flux} greatly alleviates the loss by inflating latent channels, challenges intrinsic to the DiT framework~\cite{peebles2023dit}, such as slow convergence, and high data requirements, hinder its widespread adoption in the community.

To address this, we propose a versatile three-stage \emph{\textbf{parameter gradual encoding adaption (PGEA)}} to align the SDXL UNet with the FLUX VAE. Specifically, we expand the channels in \emph{conv in} and \emph{conv out} layers (from 4 to 16) in UNet to match FLUX VAE, enabling direct modification on the SD config file to load adapted weights.
Stage \uppercase\expandafter{\romannumeral1}: we focus on distilling the knowledge from the original \emph{conv in} layer to the adapted \emph{conv in} layer. The input image is encoded through different VAEs and passed into the 4 and 16-channel \emph{conv in} layers respectively, applying a reconstruction loss between their output to update the 16-channel \emph{conv in} layer for fast adaption.
Stage \uppercase\expandafter{\romannumeral2}: we train the UNet on a standard text-to-image task for 20,000 steps, where only the 16-channel \emph{conv in} and \emph{conv out} layers are updated. 
Stage \uppercase\expandafter{\romannumeral3}: we train and update the entire UNet on the standard text-to-image task for 200,000 steps. The complete training of PGEA lasts for 4 days on $8$ H20 GPUs.
It is noteworthy that the adapted general UNet with extremely low decoding loss can be applied to various downstream tasks. 
Crucially, despite marginal gains in quantitative metrics, RAGDiffusion's breakthrough in brand logo fidelity makes it commercially viable, the first in industrial garment restoration.


%% file: sec/4_expriment.tex
 \vspace{-2mm}
\section{Experiments}
\vspace{-2mm}
\subsection{Experimental setup}
 \vspace{-2mm}
\label{sec: setup}

\begin{figure}[t]
  \centering
  \includegraphics[width=1\linewidth]{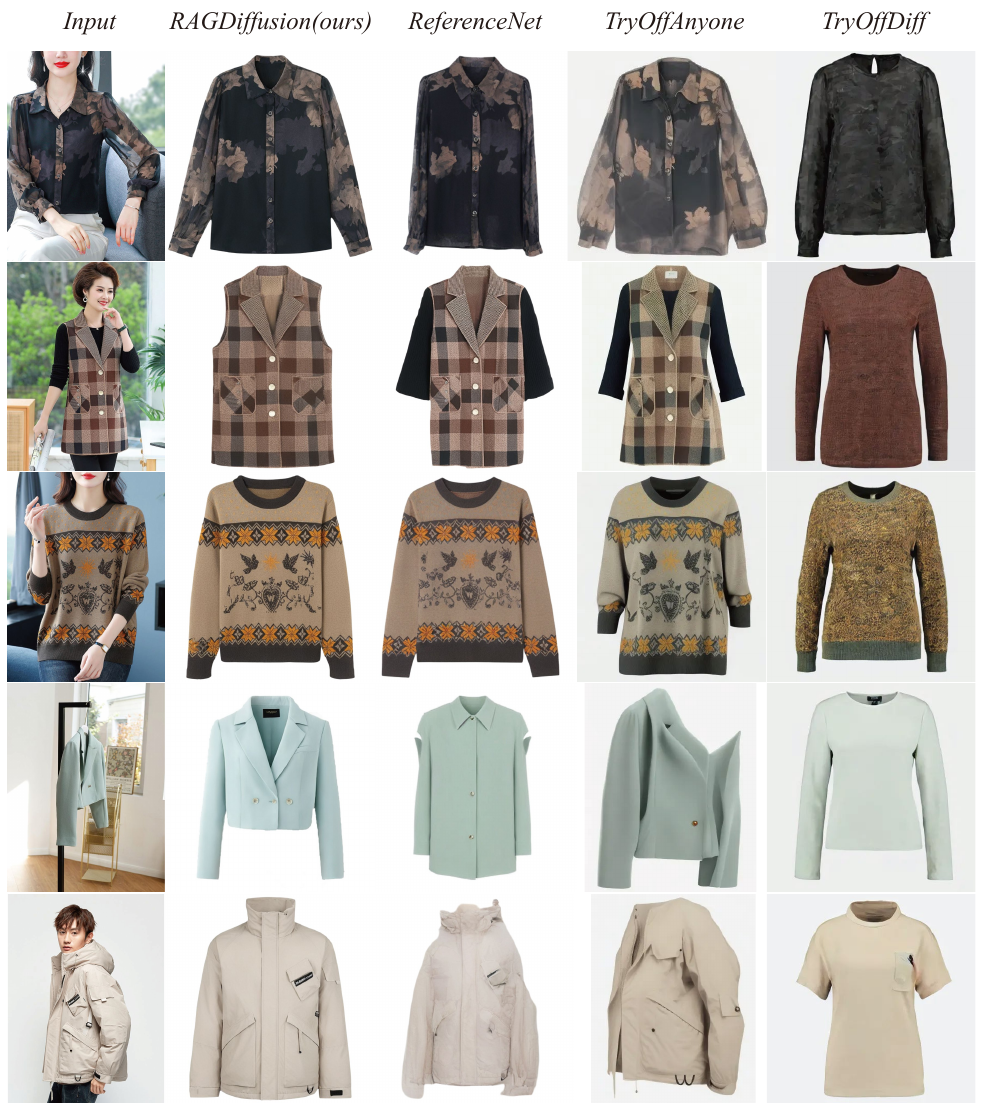}
  \vspace{-6mm}
  \caption{RAGDiffusion delivers both loyal structures and superior details on challenging layered and side-view situations.}
  \vspace{-3mm}
  \label{fig: main}
\end{figure}

\noindent \textbf{Datasets.} 
We collected 65,131 pairs of (in-the-wild upper clothing, standard flat lay upper clothing) for training and 1969 pairs for testing, named STGarment. Among the in-the-wild clothing, there are three main displays: clothing worn on a person, clothing laid indoors, and clothing hung on hangers. We use Qwen2-VL-7B~\cite{bai2023qwen} to provide prompt annotations for each pair along with 10 discrete attributes (3.5 seconds per image). Additionally, we matched each pair of images with the most structurally similar flat-lay clothing for StructureNet training based on canny image similarity. We employed a clustering approach to filter and construct a memory database for retrieval, which contains 4,000 embedding-landmark pairs from the training set. Please refer to the Appendix for more details.

\begin{table}[t]
\centering
\fontsize{7pt}{7pt}\selectfont
\setlength{\tabcolsep}{6pt}  
\begin{tabular}{lcccccc}
\toprule
\multicolumn{2}{l}{Method}       & SSIM $\uparrow$ & LPIPS $\downarrow$ & FID $\downarrow$ & KID $\downarrow$  & DISTS $\downarrow$ \\ 
\cmidrule{1-7}  
\multicolumn{2}{l}{PBE}    & 0.5278 & 0.4821  & 26.55  & 10.08 & 0.352  \\
\multicolumn{2}{l}{ControlNet}  & 0.5225 & 0.4792  & 23.02  & 9.231 & 0.345   \\
\multicolumn{2}{l}{IP-Adapter}  & 0.6067 & 0.4728  & 14.50  & 3.747 & 0.264    \\
\multicolumn{2}{l}{TryOffDiff}  & 0.6124 & 0.5007  & 34.34  & 14.19 & 0.335  \\
\multicolumn{2}{l}{TryOffAnyone}  & \underline{0.6575} & 0.4335  & 18.86  & 7.919 & 0.245     \\
\multicolumn{2}{l}{ReferenceNet}  & 0.6546 & \underline{0.3979}  & \underline{10.70}  & \underline{1.399} & \underline{0.224}    \\
\cmidrule{1-7}  
\multicolumn{2}{l}{\textbf{RAGDiffusion}}   & \textbf{0.6963} & \textbf{0.3684}  & \textbf{9.990}  & \textbf{1.092} & \textbf{0.192}   \\
\bottomrule
\end{tabular}
\vspace{-2mm}
\caption{Quantitative comparisons on the STGarment dataset.} 
\vspace{-5mm}
\label{tab: main}
\end{table}

\begin{table}[t]
\centering
\fontsize{7pt}{8pt}\selectfont
\setlength{\tabcolsep}{5pt} 
\begin{tabular}{lcccccc}
\toprule
\multicolumn{2}{l}{Method}          & SSIM $\uparrow$ & LPIPS $\downarrow$ & FID $\downarrow$ & KID $\downarrow$  & DISTS $\downarrow$ \\ 
\cmidrule{1-7}  
\multicolumn{2}{l}{Vanilla Model}    & 0.6576 & 0.3961  & 10.65  & 1.405 & 0.222     \\
\multicolumn{2}{l}{+ emb}         & 0.6614 & 0.3917  & 10.55  & 1.354 & 0.217     \\
\multicolumn{2}{l}{+ emb + sil}     & 0.6926 & 0.3690  & 10.22  & \textbf{0.940} & 0.210     \\ 
\multicolumn{2}{l}{Generating mask }     & 0.6426 & 0.4186  & 14.55  & 3.570 & 0.239    \\
\multicolumn{2}{l}{+ emb + sil + PGEA \textbf{(Full)}}   & \textbf{0.6963} & \textbf{0.3684}  & \textbf{9.990}  & 1.092  & \textbf{0.192}    \\ 
\bottomrule
\end{tabular}
\vspace{-2mm}
\caption{Quantitative ablation study of each component.} 
\vspace{-3mm}
\label{tab: ablation}
\end{table}

\begin{table}[t]
\fontsize{7pt}{7pt}\selectfont
\setlength{\tabcolsep}{4pt} 
\centering
        \begin{tabular}{l c c c c c c c c c}
          \toprule
           \multicolumn{2}{l}{Dataset} &\multicolumn{2}{c}{Viton-HD} &\multicolumn{2}{c}{DC-upper} &\multicolumn{2}{c}{DC-lower} &\multicolumn{2}{c}{DC-dress} \\
            \cmidrule{1-2} \cmidrule{3-4}  \cmidrule{5-6} \cmidrule{7-8} \cmidrule{9-10} 
          \multicolumn{2}{l}{Method}                              
          & FID & KID  & FID  & KID & FID  & KID  & FID  & KID   \\
            \cmidrule{1-2} \cmidrule{3-10} 
        \multicolumn{2}{l}{PBE} 
          & 80.8  & 52.9 & 45.0 & 21.1 & 158.8 & 128.3 & 105.1 & 74.6  \\  
          \multicolumn{2}{l}{ControlNet} 
          & 69.9 & 43.2 & 48.9 & 23.6 & 174.7 & 161.7 & 118.3 & 95.4 \\  
          \multicolumn{2}{l}{IP-Adapter} 
          & 46.8 & 17.8 & 28.8 & 11.1 & \underline{111.3} & \underline{62.7} & 47.8 & \underline{19.0} \\  
          \multicolumn{2}{l}{TryOffDiff*} 
          & \textcolor{gray}{14.0} & \textcolor{gray}{3.31} & 38.9 & 16.7 & 158.7 & 130.0 & 74.3 & 48.6 \\  
          \multicolumn{2}{l}{TryOffAnyone*} 
          & \textcolor{gray}{\textbf{11.4}} & \textcolor{gray}{\underline{1.97}} & 24.4 & 9.30 & 114.2 & 79.6 & \underline{38.4} & 20.4 \\  
          \multicolumn{2}{l}{ReferenceNet} 
          & 15.2 & 3.23 & \underline{17.8} & \underline{6.24} & 124.5 & 90.2 & 51.2 & 22.4 \\
            \cmidrule{1-2} \cmidrule{3-10}
          \multicolumn{2}{l}{\textbf{Ours}} 
           & \underline{12.3} & \textbf{1.45} & \textbf{15.9} & \textbf{4.51} & \textbf{40.5} & \textbf{19.4 }& \textbf{23.1} & \textbf{6.22} \\
          \bottomrule
        \end{tabular}
\vspace{-2mm}
  \caption{Cross dataset evaluation on Viton-HD, DressCode. Model with * means inner-dataset test which is trained on Viton-HD.}
\vspace{-3mm}
  \label{tab:merge:new_data}
\end{table}

\noindent \textbf{Implementation details.}
We initialize the RAGDiffusion with a pre-trained SDXL model and train it on STGarment using an AdamW optimizer with the learning rate of $5e-5$ at a resolution of $768 \times 768$. The models are trained for $5$ days on $8$ H20 GPUs with DeepSpeed~\cite{deepspeed} ZeRO-2 to reduce memory usage, at a batch size of $15$. $K=4$ in SLLE in \cref{eq: least_squared}. The StructureNet (\ie embedding encoder) utilizes CLIP-ViT-L/14 image encoder as the backbone and is fine-tuned on nearly 2 million various garment images following DinoV2~\cite{oquab2023dinov2}. StructureNet is further trained on STGarment with contrastive learning for $4$ days on $4$ H20 GPUs at a batch size of $128$. At inference time, we run RAGDiffusion on a single NVIDIA RTX 3090 GPU for $30$ sampling steps with the DDIM sampler~\cite{song2020ddim}. Please refer to the Appendix for more details.

\noindent \textbf{Evaluation protocols.}
About \textbf{generation quality}, we utilize LPIPS~\cite{zhang2018perceptual}, SSIM~\cite{wang2004ssim} to assess the reconstruction accuracy, DISTS~\cite{dists} to evaluate image similarity on both perceptual and structural level. Additionally, we employ FID~\cite{parmar2021cleanfid} and KID~\cite{sutherland2018kid} metrics to evaluate the realism and authenticity of the generated distributions. In terms of \textbf{retrieval ability}, we evaluate the top-1 accuracy and top-5 accuracy of retrieval results from the memory database $\mathcal{D}$ across 1969 test samples, alongside the average Intersection over Union (IoU) of the corresponding silhouette masks with GT ones. Given the absence of GT masks of test samples in the memory database $\mathcal{D}$, we define the retrieved masks to be accurate if their IoU exceeds $0.85$.

\noindent \textbf{Baselines.}
We compare RAGDiffusion with recent works TryOffDiff~\cite{velioglu2024tryoffdiff} and TryOffAnyone~\cite{xarchakos2024tryoffanyone} with official checkpoints. We also train 4 classic conditional generation methods on STGarment with SDXL backbone for fair comparison: IP-Adapter~\cite{ye2023ipadapter}, ReferenceNet~\cite{hu2023animate}, ControlNet~\cite{zhang2023controlnet} and Paint-by-Example~\cite{yang2023paintbyexample} (PBE) as baselines.

\begin{table}[t]
\centering
\fontsize{8pt}{9pt}\selectfont

\begin{tabular}{lccccc}
\toprule
 \multicolumn{2}{c}{Scale}      && Top-1 Acc. $\uparrow$ & Top-5 Acc. $\uparrow$ & IOU $\uparrow$  \\ 
\cmidrule{1-2} \cmidrule{4-6} 
\multicolumn{2}{c}{1000}   && 76.6\%  & 93.4\%  & 0.872    \\
\multicolumn{2}{c}{2000}   && 78.2\%  & 93.2\%  & 0.885    \\ 
\multicolumn{2}{c}{4000}   && \textbf{80.6\%}  & \textbf{93.7\%}  & \underline{0.902}   \\
\multicolumn{2}{c}{8000}   && \underline{79.7\%}  & \underline{93.4\%}  & \textbf{0.905}   \\
\bottomrule
\end{tabular}
\vspace{-2mm}
\caption{Retrieval accuracy at different scales of external database.}
\vspace{-5mm}
\label{tab: retrieval}
\end{table}

\begin{figure}[t]
  \centering
  \includegraphics[width=1\linewidth]{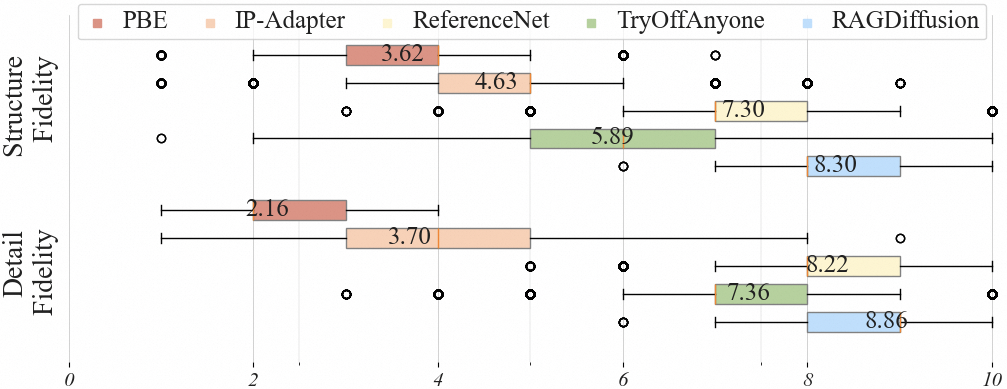}
  \vspace{-7mm}
  \caption{Boxplot illustration of user study. RAGDiffusion demonstrates better performance and stability across scenes.}
  \vspace{-6mm}
  \label{fig: user_study}
\end{figure}

\subsection{Generation comparisons with baselines}
 \vspace{-2mm}
\noindent \textbf{Qualitative results.}
\cref{fig: main} presents a qualitative comparison between RAGDiffusion and baselines on the STGarment dataset. TryOffDiff has yielded entirely unsuccessful outputs. Meanwhile, naive ReferenceNet and TryOffAnyone manage to maintain correct textures in challenging scenarios, but encounter structural confusion and distortion. This phenomenon may stem from TryOffDiff and ReferenceNet's over-reliance on local visual texture features, lacking the benefits of a global perspective in challenging cases. RAGDiffusion employs a retrieval-aggregate approach to capture structural information and integrates conditional controls from EP-Adapter, Landmark Guider and PGEA, yielding promising performance in both structurally faithful and detail-oriented garment conversion. Notably, while trained on the same dataset with naive ReferenceNet, RAGDiffusion assimilates high-quality contour landmarks and structure embeddings as external prior to producing visually compelling results that enhance depth and realism.

\begin{figure*}[t]
\centering
\begin{minipage}[t]{0.685\linewidth}
\centering
\includegraphics[width=1\linewidth]{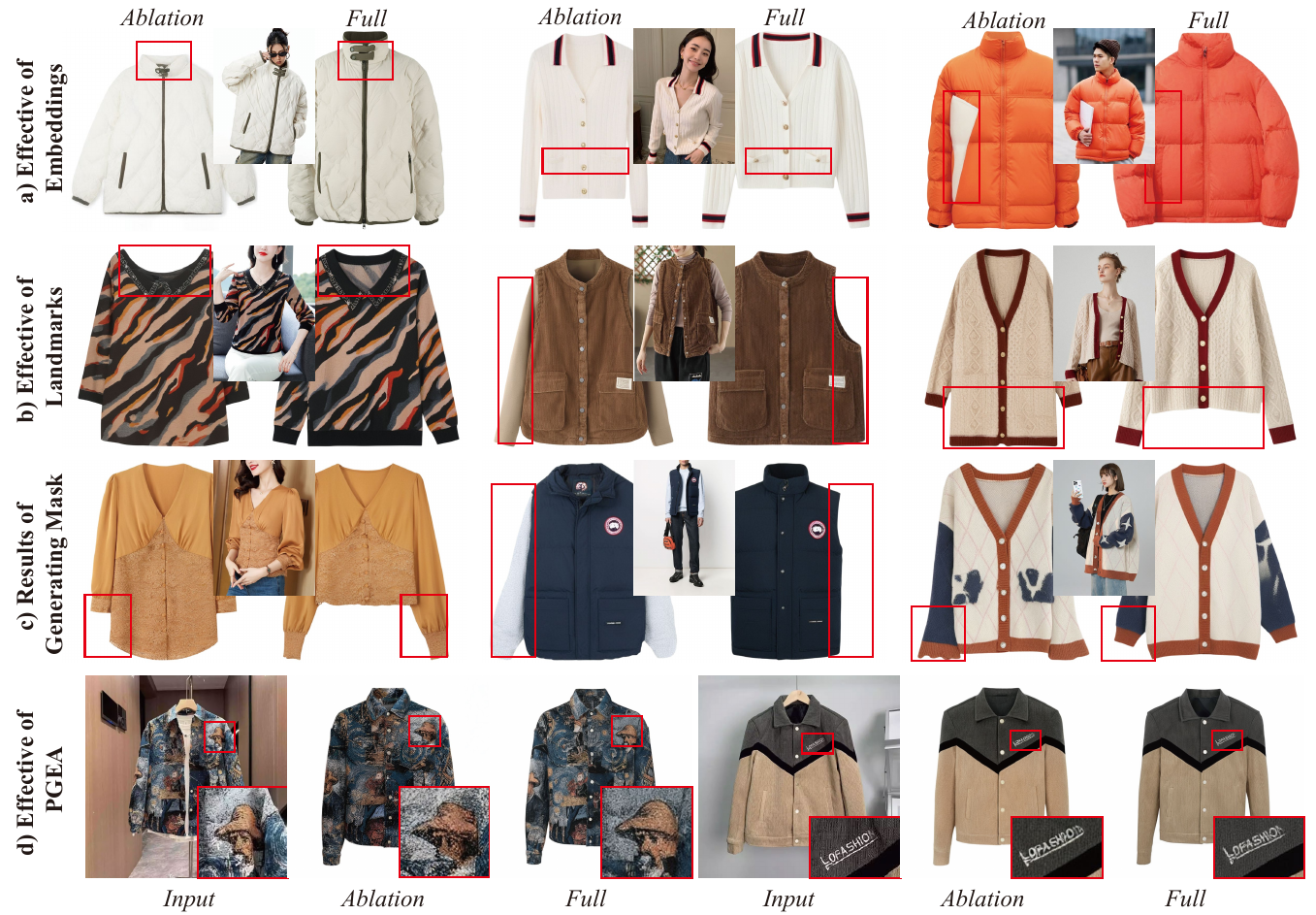}
\vspace{-6mm}
\caption{Ablation study. Without landmarks, RAGDiffusion suffers inaccurate shape. 
  Embeddings from StructureNet improve inner structure, while PGEA enhances detail preservation.}
\label{fig: ablation}
\end{minipage}
\begin{minipage}[t]{0.295\linewidth}
\centering
\includegraphics[width=\linewidth]{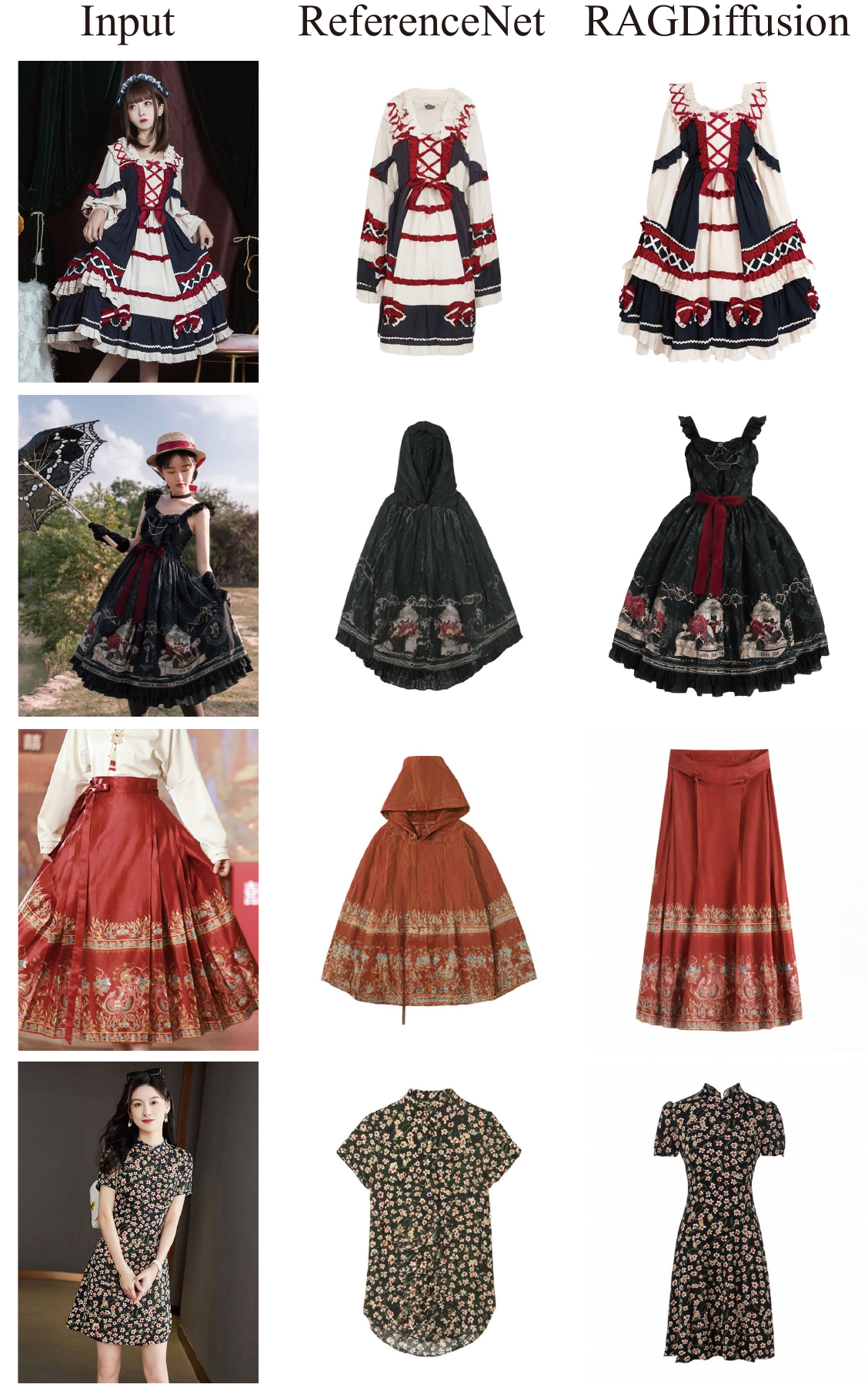}
\vspace{-6mm}
\caption{RAG significantly improves generalization on untrained categories.}
\label{fig: lower_general}
\end{minipage}
\vspace{-5mm}
\end{figure*}

\noindent \textbf{Quantitative results.}
As indicated in \cref{tab: main}, we quantitatively evaluate the \emph{generation quality} of various methods on the STGarment dataset, demonstrating that RAGDiffusion significantly outperforms all baseline approaches. This confirms RAGDiffusion's ability to deliver superior and accurate garment generation across diverse scenarios. 
Retrieval, as a crucial component of RAGDiffusion for assimilating external knowledge, plays a significant role. Using the evaluation metrics described in \cref{sec: setup}, we report the \emph{retrieval accuracy} at different sizes of the external memory database $\mathcal{D}$ in \cref{tab: retrieval}. Specifically, we employ clustering and downsampling algorithms to iteratively eliminate outliers and highly similar samples from the original memory database, constructing retrieval libraries of four different scales: 1000, 2000, 4000, and 8000 samples. 
When the memory database $\mathcal{D}$ is too large, outliers may adversely affect quality if embeddings are inaccurate, while redundant samples reduce retrieval efficiency. Conversely, if the memory database $\mathcal{D}$ is too small, the retrieval library may lack sufficient representativeness and completeness, adversely affecting the generation performance for specific categories. Notably, a memory database comprising 4000 samples achieved the best trade-off in performance. Please refer to the Appendix for more technical details.

\noindent \textbf{User study.} 
When metrics like FID assess the realism and authenticity of standard garment generation, they tend to be less sensitive to high-frequency information such as contours, collar variations, buttons, and color shifts~\cite{chen2024zero}. We conducted user studies involving 100 participants to evaluate different methods based on user preferences across 200 randomly selected image pairs from the test set. Participants were asked to assign a preference score (ranging from $1 \sim 10$) in terms of structural fidelity and detail fidelity for each sample generated by anonymized methods. As illustrated in \cref{fig: user_study}, we report the distribution of the scores, encompassing medians, means, quartiles, and outliers. Our findings reveal that RAGDiffusion is significantly preferred over all baseline methods in terms of structural fidelity ($mean=8.34$) and detail fidelity ($mean=8.52$). Additionally, the narrow range of outliers in our method indicates a more stable generation across various scenes.

\vspace{-2mm}
\subsection{Ablation study}
\label{sec: ablation}
 \vspace{-2mm}
 To validate our core contributions, we define a Vanilla Model that removes structure embedding $\hat{e}_{itw}$, silhouette landmarks $\hat{L}_{sil}$ and PGEA. Specifically, its $\hat{e}_{itw}$ and $\hat{L}_{sil}$ are set to zero vectors while using original SDXL VAE.
 

\noindent \textbf{Soft guidance of structure embeddings.}
Since structure embeddings serve as prerequisites for retrieval in SLLE and inputs to EP-Adapter, we first add structure embeddings $\hat{e}_{itw}$ into EP-Adapter of vanilla model (denoted as ``+ emb''). As in \cref{tab: ablation} and \cref{fig: ablation}, vanilla model lacking structure embedding frequently exhibits artifacts in clothing internal structures, such as inaccurate collars, missing pockets, and occlusion confusion. This demonstrates that structure embedding provides global guidance and plays a fundamental role in optimizing apparel internal structures.

\noindent \textbf{Hard guidance of silhouette landmarks.}  
The core of RAG is to retrieve silhouette landmarks that enhance explicit spatial constraints. Thus we add landmarks on ``+ emb'' version as the ablation named ``+ emb + sil'' in \cref{tab: ablation}. The model without retrieval exhibits significant contour errors, particularly around the back collar, sleeve length, and garment length. This is attributed to the fact that retrieval-augmented SLLE incorporates priors from LLM, which transcend visual limitations and aid in accurate spatial structures through the use of landmarks. Furthermore, we investigate the role of SLLE in embedding remapping in \cref{sec: analysis}.

\noindent \textbf{Alternative to RAG: generating masks via UNet.}
As RAG requires a memory database and contrastive learning for embeddings, we try a simpler alternative by employing an SDXL UNet to directly predict landmarks. Specifically, we modify the input of the alternative UNet to a three-channel in-the-wild image $x_{itw}$, with outputs being one-channel predicted landmarks $L_{sil}$ that are directly fed into RAGDiffusion's denoising UNet. However, this alternative produces landmarks with noticeable errors and degraded generation quality, due to limitations in visual understanding and the absence of proper pre-trained weights.

\noindent \textbf{PGEA.} 
PGEA is employed to mitigate information loss during the VAE encoding and decoding process. We add PGEA on ``+ emb + sil'' version as Full Model. The full model utilizing PGEA achieves clearer and more accurate edges in high-frequency patterns simulation; additionally, it shows significant improvements in the recovery of logos and text. PGEA greatly alleviates the long-standing issue of distortion in detail and makes RAGDiffusion commercially viable, the first in industrial garment restoration.

\subsection{Additional rationale for introducing RAG}
\vspace{-2mm}
In addition to structural faithfulness, RAGDiffusion demonstrates \textbf{excellent generalizability and robustness} across untrained categories and scenes simply by updating retrieval database, without the need for collecting expensive paired data and retraining. Besides, please refer to Appendix for another rationale of \textbf{human-interpretable control}.

\noindent \textbf{Cross category evaluation.} RAGDiffusion is trained on upper-body clothing data and has not encountered lower-body garments during training. To validate the generalizability,  we collect $856$ flat-lay lower-body/dress images (no in-the-wild image is needed) and incorporate them with silhouette masks into the external memory database through StructureNet. Subsequently, we test it on $50$ lower-body in-the-wild clothing samples, using a ReferenceNet as baseline. The results in~\cref{fig: lower_general} demonstrate that retrieval significantly improves generalization capabilities, serving as a cost-effective zero-shot generalizing solution. The quantitative results on lower-body/dress categories in~\cref{tab:merge:new_data} also validate the generalizability boost. 

\noindent \textbf{Cross dataset evaluation.}
Despite zero exposure to the Viton-HD~\cite{choi2021vitonhd} and DressCode~\cite{he2024dresscode} datasets during training, RAGDiffusion shows strong out-of-distribution (OOD) compatibility over baselines without tuning illustrated in ~\cref{tab:merge:new_data}. This boost in generalization increases the operational maturity of RAGDiffusion, enabling it to effectively handle various OOD images submitted by users.

\vspace{-2mm}
\subsection{Component analysis}
\label{sec: analysis}
 \vspace{-2mm}
 
\begin{figure}[t]
  \centering
  \includegraphics[width=1\linewidth]{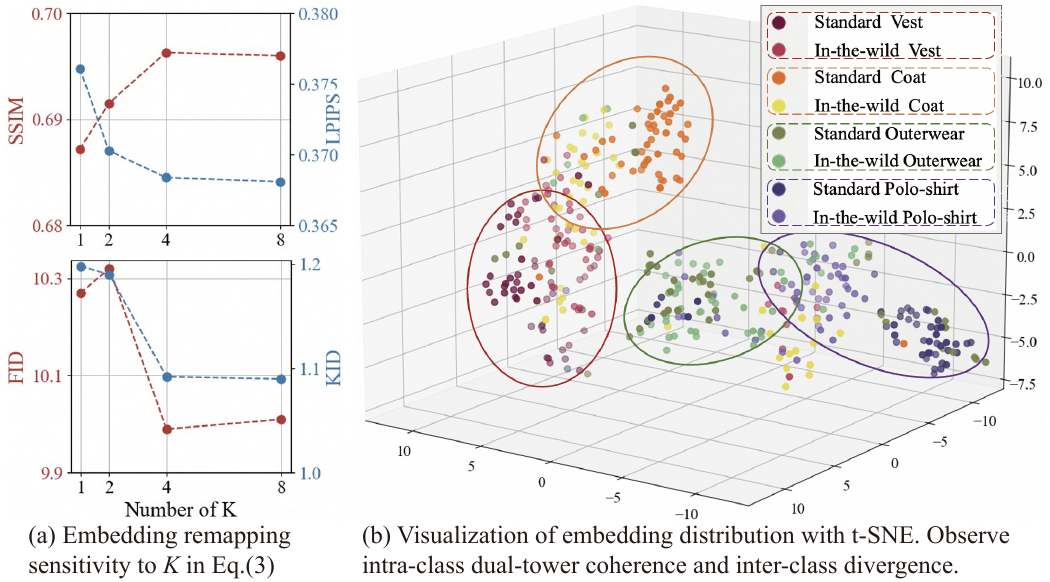}
  \vspace{-6mm}
  \caption{Illustrations of component analysis.}
  \label{fig: tsne}
  \vspace{-3mm}
\end{figure}

\begin{figure}[t]
  \centering
  \includegraphics[width=1\linewidth]{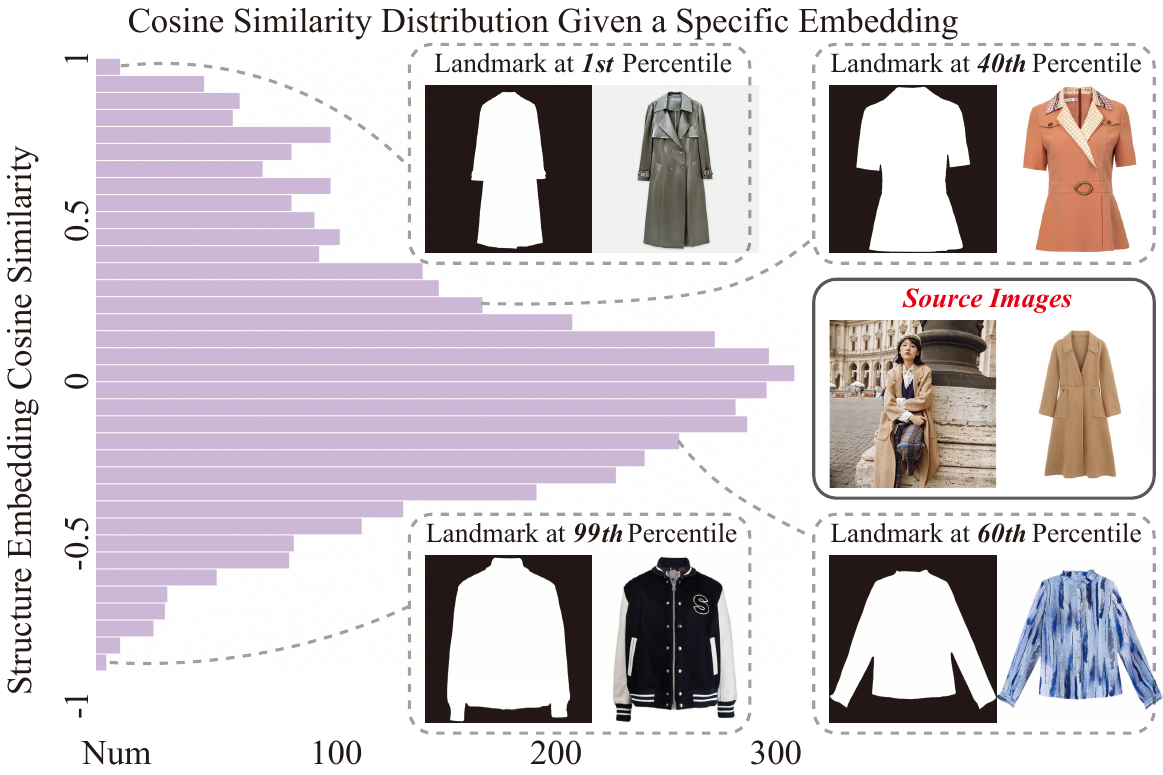}
  \vspace{-6mm}
  \caption{Distribution of cosine similarity between a given sample and images from the external memory database.}
  \vspace{-5mm}
  \label{fig: rank}
\end{figure}

\noindent \textbf{Latent structure embedding distributions.}
We further visualize the learned latent embedding distribution in \cref{fig: tsne}(b) to gain an in-depth comprehension of how the embeddings work in retrieval and the EP-Adapter. For this purpose, we sample $N=200$ pairs of (in-the-wild cloth embedding $e_{itw}$, standard cloth embedding $e_{std}$) pairs from our dual-tower StructureNet. We then employ t-SNE~\cite{van2008visualizing} to project each feature representation into a point within \cref{fig: tsne}. Notably, 1) the dual-branch embeddings ($e_{itw}$, $e_{std}$) corresponding to the same category exhibit clustering behavior, indicating that the learned priors in StructureNet effectively perform structural alignment cross domains, thus facilitating retrieval based on structural similarity; and 2) representations from different categories are clearly dispersed, demonstrating that the embeddings encompass discriminative structural information, which aids in generation through the EP-Adapter. Lastly, we also visualize the distribution of cosine similarity between a given sample and images from the external memory database $\mathcal{D}$ in \cref{fig: rank}. The cosine similarities present a normal distribution, affirming that the constructed embedding library is representative and comprehensive. Furthermore, the examples visualized for different similarity retrieval results effectively illustrate the efficacy of the landmark retrieval mechanism.

\noindent \textbf{Number of retrieved embeddings.}
As the number $K$ of retrieved nearest neighbors in \cref{eq: least_squared} during SLLE plays a fundamental role in the properties of the final structure embeddings and landmarks, we demonstrate the fusion sensitivity to $K$ in \cref{{fig: tsne}}(a). A smaller $K$ value facilitates the sharp edge of landmarks, whereas a larger $K$ value enhances the accurate representation of the reconstructed embedding $e_{itw}'$ in \cref{eq: fuse}. It can be observed that the optimal performance occurs at $K=4$, which also serves as the default parameter setting for other experimental parts.

%% file: sec/5_conclusion.tex
\vspace{-1mm}
\section{Conclusion}
\vspace{-2mm}
In conclusion, RAGDiffusion presents a significant advancement in the generation of standardized clothing assets by effectively addressing the prevalent challenges of structural hallucinations and texture fidelity. By integrating retrieval-based structure aggregation to get soft and hard guidance, as well as omni-level faithful generation pipeline, RAGDiffusion is capable of synthesizing both structure and texture faithful clothing assets. Comprehensive experiments and in-depth analysis demonstrate notable performance improvements over existing models. We hope that RAGDiffusion helps open avenues for RAG in diverse high-specification faithful generation tasks, bringing us closer to the goal of professional content creation for all.

\clearpage  

\section*{Acknowledgements} 
This work is supported by the Science and Technology Commission of Shanghai Municipality under research grant No. 25ZR1401187.

%% file: sec/X_supp.tex
\clearpage
\setcounter{page}{1}
\maketitlesupplementary
In \cref{supp: sec: rationale}, we demonstrate two additional benefits brought by the RAG system: enhanced generalizability, and human-interpretable control through landmark manipulation.
In~\cref{supp: sec: limit}, we discuss the limitations of our work and outline directions for future research. 
Given that standard garment generation is a novel task, we elaborate on its practical significance in \cref{supp: sec: downstream}. 
To enhance reproducibility, we provide a detailed description of the network architecture, hyperparameters, and procedural steps in \cref{supp: sec: implementation}, thoroughly covering the process of RAG system construction, including dataset. 
In \cref{supp: sec: preliminary}, we discuss preliminary knowledge relevant to this work. 
Finally, in \cref{supp: sec: experiment}, we detail the experimental evaluation process and provide additional results from the ablation study, along with extensive cases of RAGDiffusion within STGarment, Viton-HD and DressCode.

\begin{figure}[t]
  \centering

  \includegraphics[width=1\linewidth]{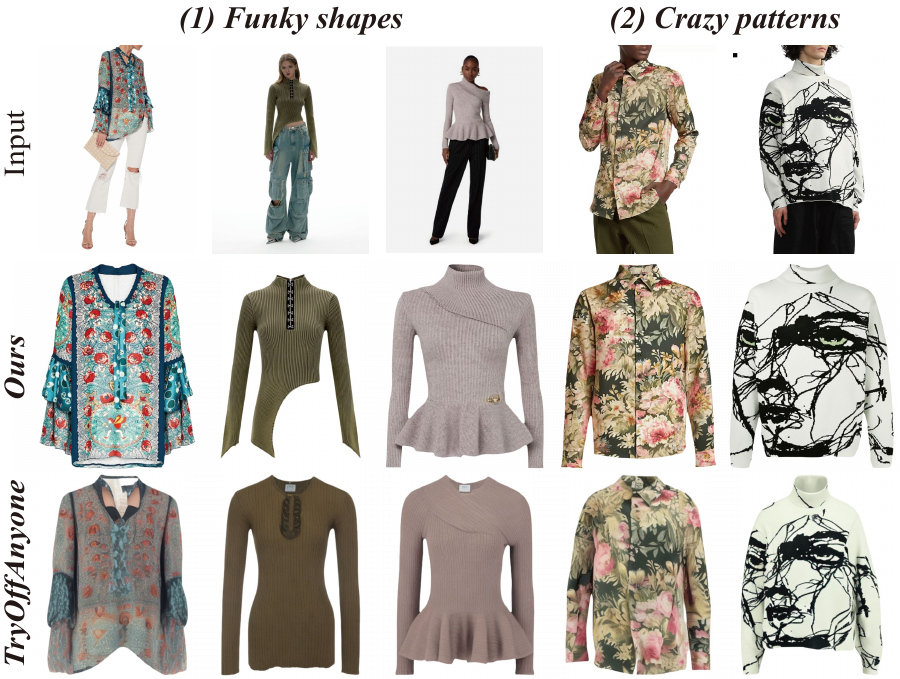}
  \caption{Generalization comparison on crazy difficult garments.}
  \label{fig: reb}
\end{figure}

\begin{figure*}[ht]
  \centering
  \includegraphics[width=1\linewidth]{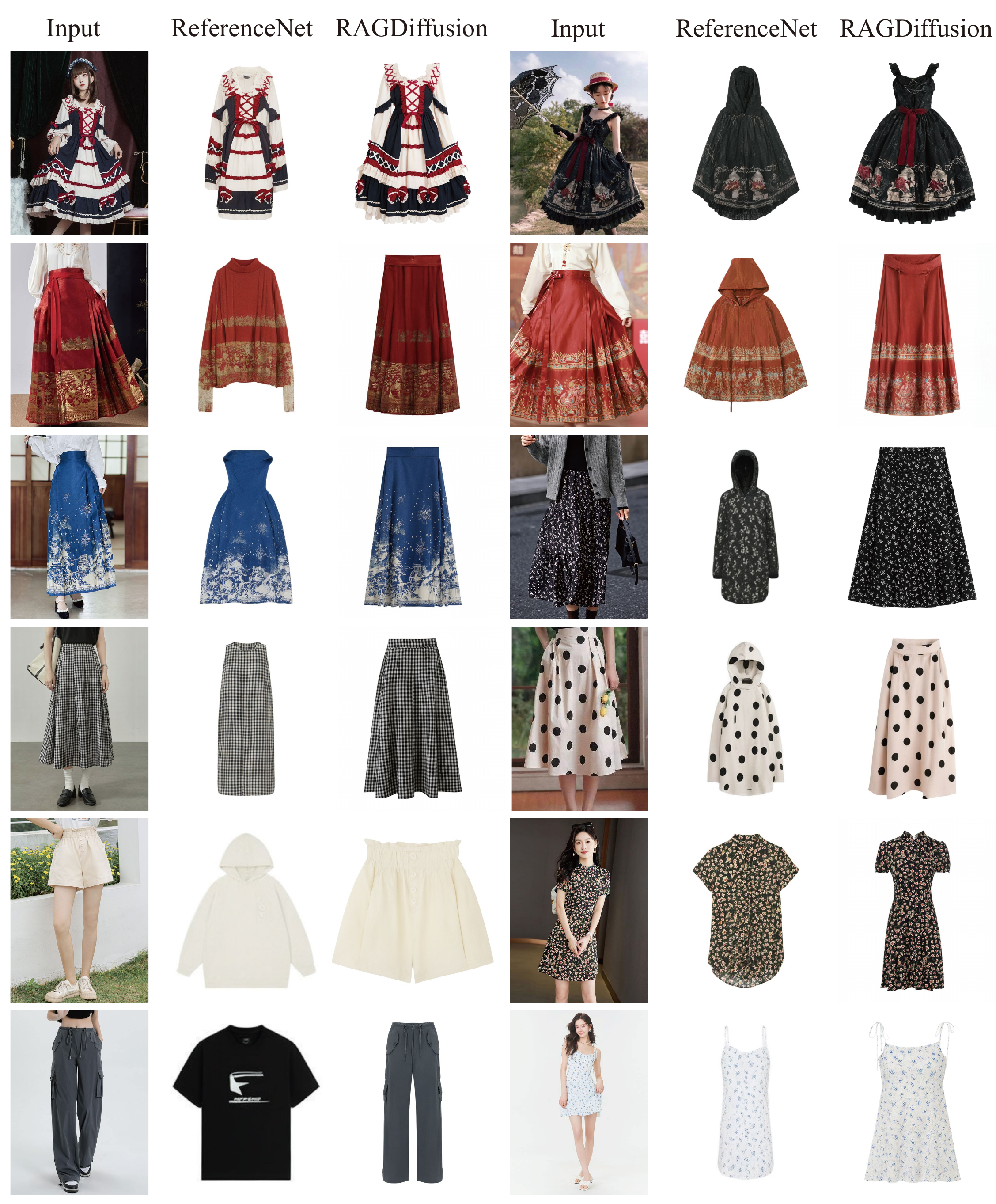}
  \caption{RAGDiffusion is able to produce accurate results in the lower-body domain, showcasing its strong out-of-distribution compatibility and generalization ability.}
  \label{supp: fig: take place for lower_body}
\end{figure*}

\begin{figure*}[ht]
  \centering
  \includegraphics[width=1\linewidth]{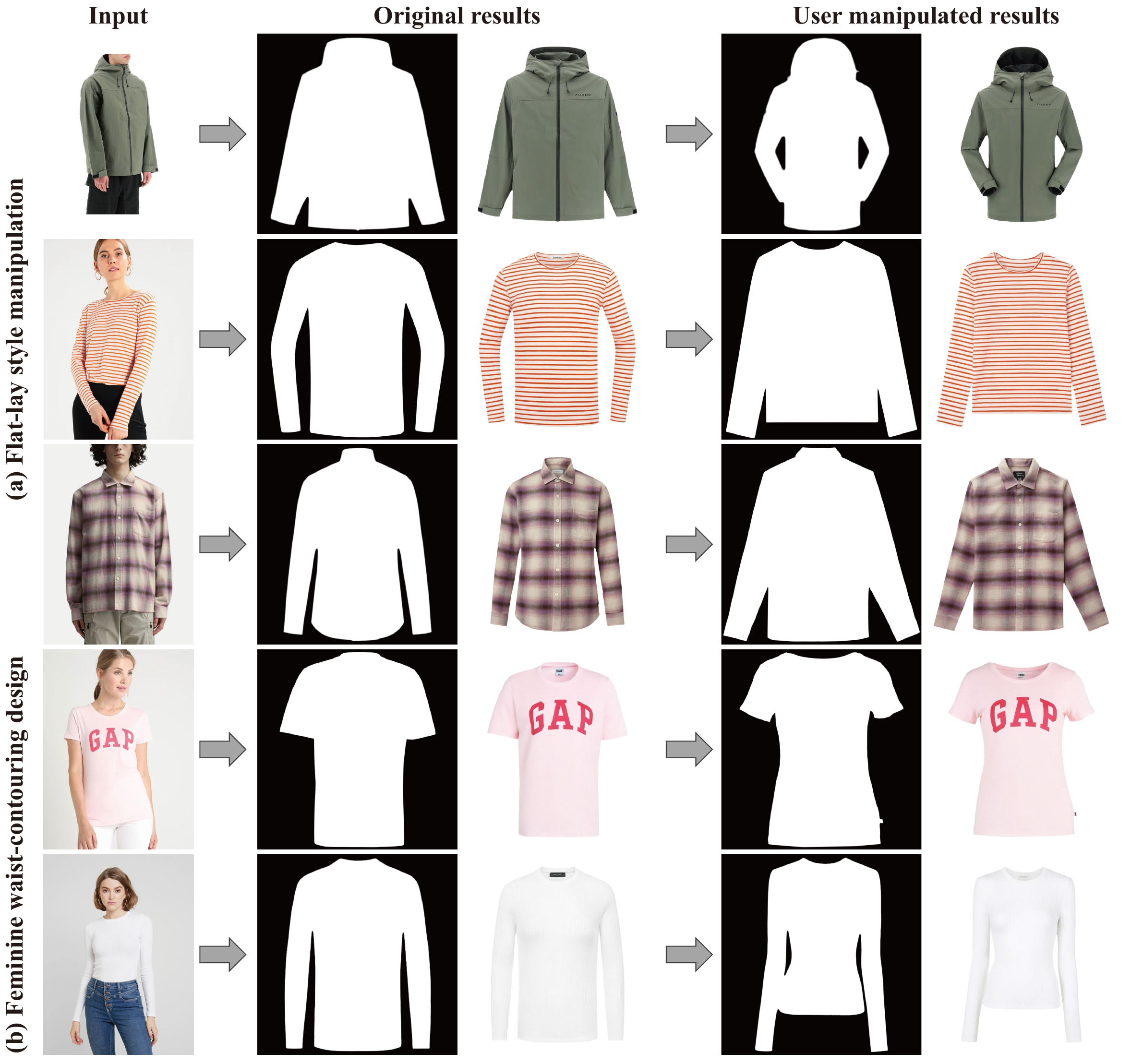}
  \caption{Users can modify the final visual presentation by replacing recommended silhouette masks/landmarks with one of $K$ candidates before UNet denoising, highlighting RAGDiffusion's advantages in human-interpretable control and retrieval-based recommendation manipulation.}
  \label{supp: fig: usermodified}
\end{figure*}

 \section{Additional rationale for introducing RAG}
 \label{supp: sec: rationale}
In addition to enhancing structural determinacy and eliminating structural distortion through the assimilation of structural landmarks and external knowledge, we have also discovered two additional benefits brought by the RAG system: enhanced generalizability, and human-interpretable control through landmark manipulation.

\subsection{Enhanced generalizability due to RAG}
Generally speaking, Retrieval-Augmented Generation (RAG) can \textbf{significantly enhance generalization and robustness in new scenarios simply by updating the retrieval database}, without the need for retraining. This provides a cost-effective and convenient maintenance solution for large generative models, avoiding the expensive process of retraining. We have also observed similar phenomena in RAGDiffusion.

RAGDiffusion is trained on upper-body clothing data and has not encountered lower-body garments during training. In this testing phase, we collect embeddings and corresponding landmarks for $856$ lower-body/dress items and incorporate them into the external memory database. Subsequently, we gather $50$ lower-body/dress in-the-wild clothing samples as a test set, using a ReferenceNet version as a baseline for comparison. The results in~\cref{supp: fig: take place for lower_body} demonstrate that retrieval significantly improves generalization capabilities.

By injecting the embeddings of lower-body garments along with their corresponding contour landmarks as conditional constraints, \textbf{RAGDiffusion produces accurate results in the lower-body domain, showcasing its strong out-of-distribution (OOD) compatibility}. In contrast, the ReferenceNet version is noticeably confused by the concept of lower-body garments and fails to yield meaningful garment structures. This is because current generative models necessitate corresponding training data to perform well. Other baselines face similar issues. \textit{ReferenceNet requires retraining} with lower-body try-on training data to work effectively on lower garments. 
In contrast, \textit{RAGDiffusion does not need retraining} or expensive try-on training data, as it can achieve results with just standard clothing database update, illustrating its greater flexibility. Of course, RAGDiffusion could perform better on lower garments if it underwent training similar to that on upper garments. This boost in generalization increases the operational maturity of RAGDiffusion, enabling it to effectively handle various OOD images submitted by users.

\noindent
\textbf{Generalization on Extreme Clothing Types.}
As RAGDiffusion works with real-world data, we provide more results to showcase how it handles super varied garments—like crazy patterns or funky designs in Fig.~\ref{fig: reb}. 
\uppercase\expandafter{\romannumeral1}.\textbf{\textit{ Crazy patterns.}} Due to the pattern-level and detail-level faithful generation pipeline (Section 3), RAGDiffusion delivers accurate texture and logos even if the garment has crazy patterns, as shown in Main. Fig. 5, 6, and Fig.~\ref{fig: reb}. 
\uppercase\expandafter{\romannumeral2}. \textbf{\textit{Funky shapes.}} Highly varied shapes pose a significant challenge for all methods. Existing methods are nearly impossible to succeed in these scenarios. While our model still has a probability of generating distorted shapes, it also produces correct results sometimes, and overall, RAGDiffusion significantly outperforms the baselines.

\subsection{Human-interpretable control through landmark manipulation due to RAG}
\label{supp: sec: manipulation}
In practice, the retrieval-acquired silhouette mask provides users with a visual shape preview opportunity before UNet denoising. Users can modify the final visual presentation by replacing recommended silhouette masks/landmarks (\ie, selecting from the recommended $K$ nearest neighbor landmark candidates during retrieval) before UNet denoising. For instance, in formal \emph{shirt} scenarios, complete flattening may be required to convey a serious aesthetic, whereas in \emph{hoodie} cases, users might prefer slightly bent sleeves with mild surface wrinkles to achieve a casual and relaxed appearance. As shown in ~\cref{supp: fig: usermodified}, we demonstrate several style modification cases of flat-lay garments through human-interpretable manipulation. This functionality is absent in end-to-end generative models like ReferenceNet~\cite{hu2023animate}, TryOffDiff~\cite{velioglu2024tryoffdiff}, and TryOffAnyone~\cite{xarchakos2024tryoffanyone}, highlighting RAGDiffusion's advantages in human-interpretable control and retrieval-based recommendation manipulation.

\section{Limitations and future work}
\label{supp: sec: limit}
We present RAGDiffusion, an efficient RAG framework that supports the generation of standardized clothing assets by effectively addressing the prevalent challenges of structural hallucinations and fidelity in generated images. However, there is also limitations to discuss, which will help us improve the proposed framework further.

\begin{figure}[t]
  \centering
  \includegraphics[width=1\linewidth]{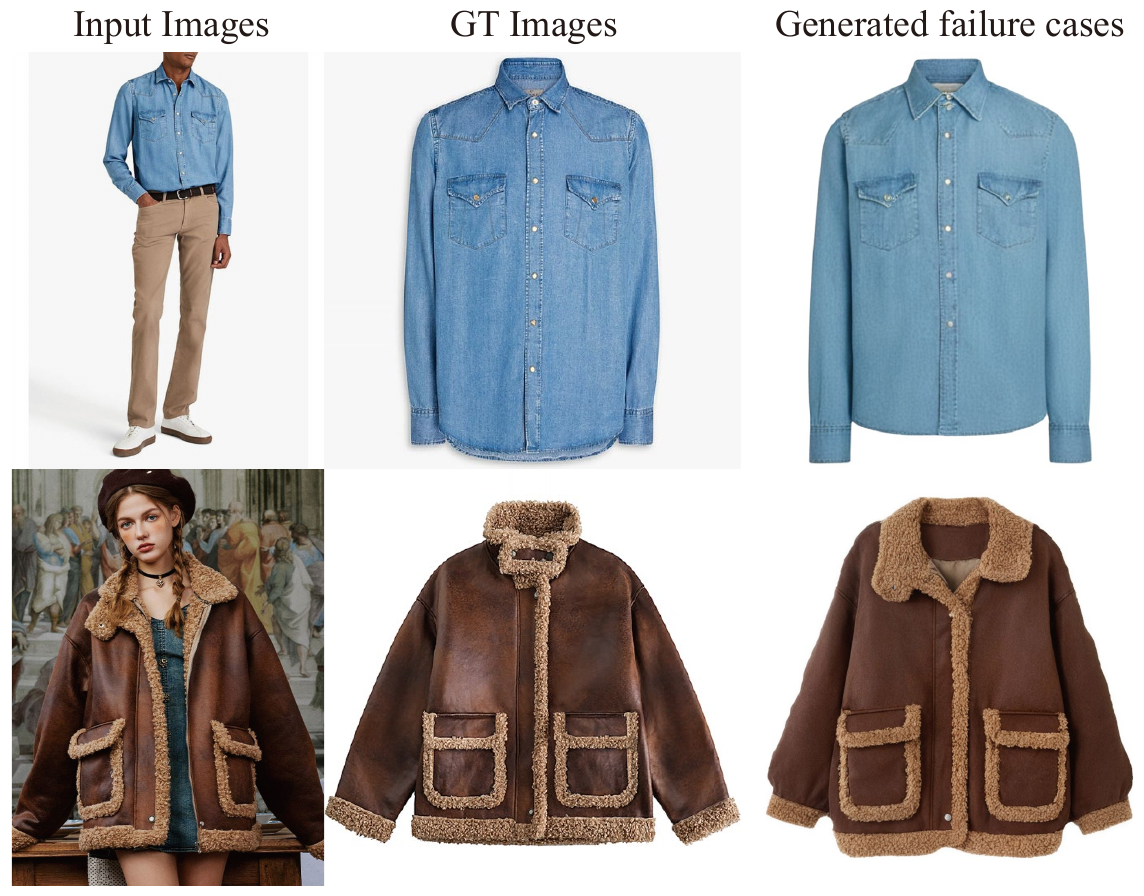}
  \caption{Failure cases about color bias. RAGDiffusion is possible to encounter color bias due to MSE loss constraints and illumination variance. Extended training durations, along with the incorporation of contrast-enhancing data augmentation techniques, can partially alleviate this issue.}
  \label{supp: fig: failcase}
\end{figure}

\begin{figure*}[t]
  \centering
  \includegraphics[width=1\linewidth]{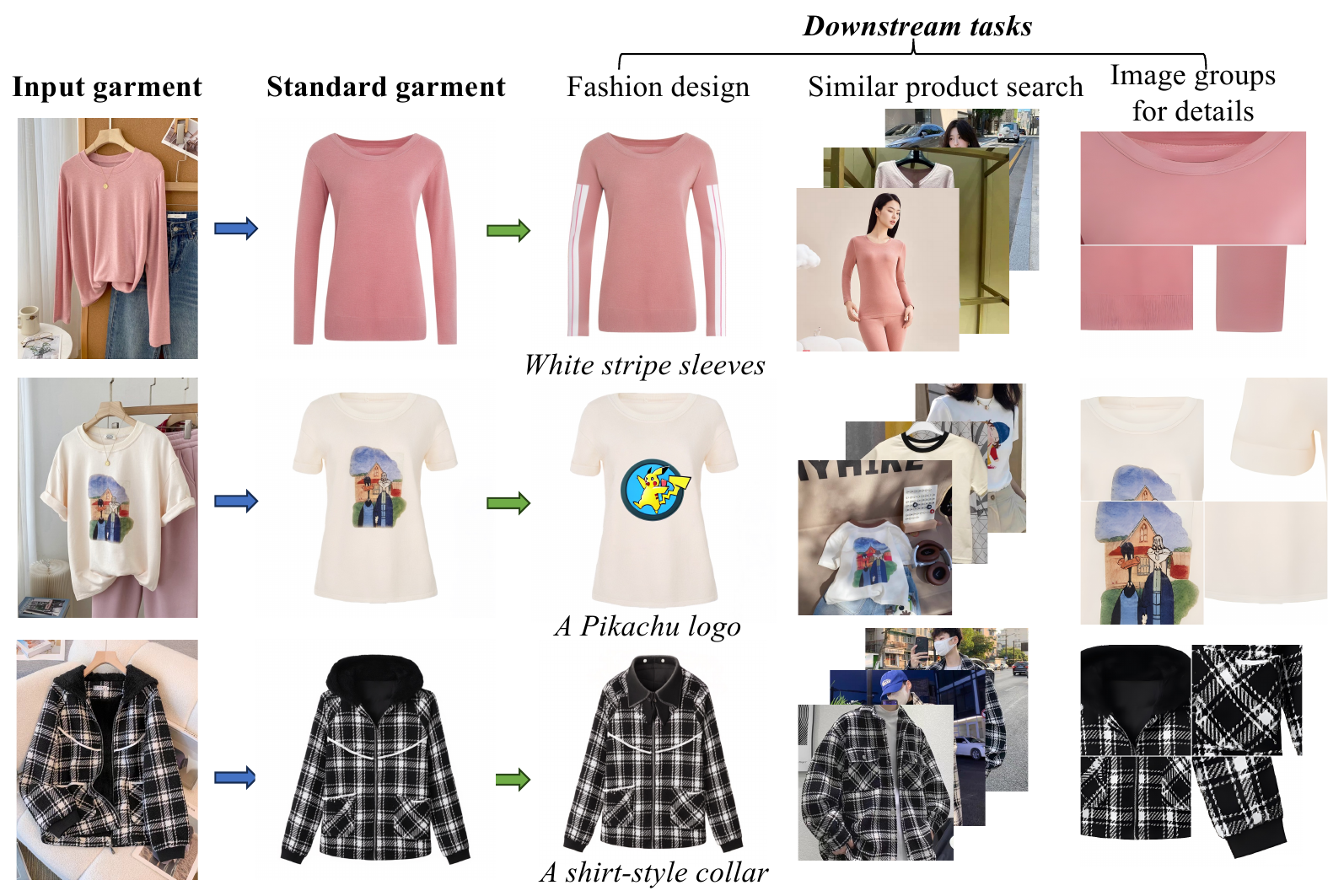}
  \caption{The standard garment serves as an essential standard element within e-commerce databases. In this context, we illustrate several examples that showcase the associated downstream applications.}
  \label{supp: fig: motivation}
\end{figure*}

Actually, we have observed \emph{a possibility of color bias in nearly pure-colored garments}, particularly under very bright or very dark situations, as shown in~\cref{supp: fig: failcase}. This phenomenon is commonly encountered in image-based image editing, as the reconstruction loss constraints of Stable Diffusion \emph{are not particularly sensitive to color discrepancies}. Additionally, \emph{the illumination variance} can further impact the perceived colors of clothing. We note that extended training durations, along with the incorporation of contrast-enhancing data augmentation techniques, can partially alleviate this issue.

In our future work, we aim to enhance RAGDiffusion in both its application depth and breadth. Firstly, by strengthening the injection of color information and incorporating lighting simulation, we hope to address the potential color bias observed in garments. Secondly, we intend to expand RAGDiffusion to encompass additional categories such as bottoms, dresses, and shoes, thereby achieving more comprehensive coverage.

\section{Downstream applications of the standard garments}
\label{supp: sec: downstream}

The standard garment acts as a vital intermediary variable connecting a range of downstream applications, including garment design, product display, and virtual fitting. It actually serves as an essential standard element within e-commerce databases. In this context, we illustrate several examples that showcase the relationships between standard garments and their associated downstream applications in~\cref{supp: fig: motivation}. The image editing results are sourced from SDXL-inpainting~\cite{podell2023sdxl}, while the remaining outcomes are derived from existing toolkits.

\begin{figure}[t]
  \centering
  \includegraphics[width=1\linewidth]{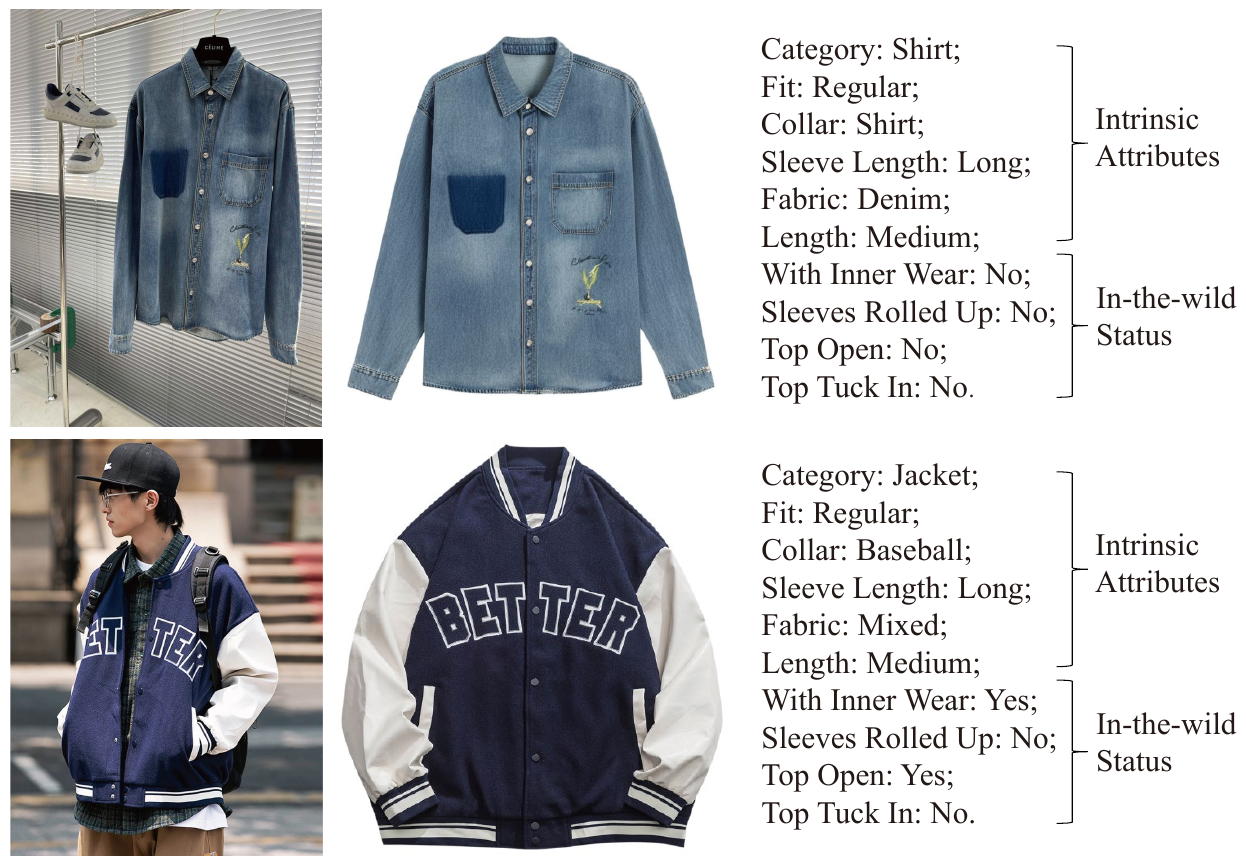}
  \caption{Visualization of data items in our collected dataset STGarment.}
  \label{supp: fig: datasetshow}
\end{figure}

\begin{table}
  \centering

  \resizebox{\linewidth}{!}{
  \begin{tabular}{lccc}
  \toprule
 Retrieval Time       & Denoising Time        & LLM Caption Time     \\
\midrule
 0.3 seconds          & 8.0 seconds         & 3.5 seconds                \\

\bottomrule
  \end{tabular}}
  \caption{Sampling time cost on an RTX 3090 GPU at resolution of $768\times 768$. Retrieval cost is a negligible part of generation.}
  \label{tab: time}
\end{table}

\section{More implementation details}
\label{supp: sec: implementation}

\subsection{Computational costs report}
The computational overhead is relatively negligible when facing a larger retrieval memory database. 
In practical generation, we store feature vectors of a standard garment set locally as the retrieval memory database. During sampling, we retrieve the most similar garment vectors efficiently with FAISS~\cite{douze2024faiss} from the memory database in parallel based on the feature vector of the input image. On the other hand, the speed bottleneck of the generation process lies in the iterative denoising process, where retrieval time is a negligible part, as shown in Tab.~\ref{tab: time}.

\subsection{StructureNet}

The extraction of pure visual features from in-the-wild clothing images presents challenges due to several local factors, including but not limited to: 1) occlusions or creases resulting from complex poses that obscure clothing content; 2) the shapes of garments being less pronounced when worn on a person, making it difficult to assess fit accuracy; and 3) in-the-wild images often containing foreground obstructions or inner garments, which can cause visual generative models to mistakenly incorporate these external objects into the final generation results. To mitigate the limitations of pure visual feature extraction, we introduce an LLM that leverages extensive pre-training on vast amounts of data to enhance our understanding of clothing.

\paragraph{Discrete attributes prediction by LLM.}
As shown in \cref{appendix: tab: attri}, we employ the Qwen2-VL-7B~\cite{bai2023qwen} language model to extract a total of $14$ clothing attributes, of which $10$ contribute to the construction of the garment structure embeddings, indicated by an asterisk $*$. The language model also provides captions describing the clothing content, thereby enhancing our external knowledge. We utilize $4$ NVIDIA H20 GPUs to deploy the Qwen2-VL-7B service, achieving the attribute and caption generation time of under 2 seconds.

\paragraph{Embedding encoding.}
In the construction of StructureNet, we first extract features \( f_{img} \) from in-the-wild clothing images or flat-lay images using the CLIP-ViT-L/14~\cite{radford2021clip} image encoder backbone (specifically from the \emph{second-to-last} layer), resulting in a feature dimension of 768. Subsequently, we assign a 32-dimensional learnable embedding to each attribute extracted from the language model and concatenate the attribute features \( f_{attr} \) with the image features \( f_{img} \) to form a vector in $\mathbb{R}^{768+10\times 32}$. Through the Resampler module~\cite{ye2023ipadapter}, we conduct a nonlinear mapping, ultimately forming an embedding \( e \). The Resampler module consists of 4 layers of MLP and 4 layers of attention mechanisms, derived from the IP-Adapter open-source code. For features originating from different image domains \( (x_{itw}, x_{std}) \), we designate the extracted embeddings as \( (e_{itw}, e_{std}) \). The dual-tower ViT image encoder combined with the Resampler module is referred to as StructureNet. As described in the main body of paper, StructureNet is trained on STGarment using contrastive learning for 4 days on 4 NVIDIA H20 GPUs with a batch size of 128.

\begin{table*}[h]
\centering
\caption{All attributes and their values are displayed. The total of $14$ clothing attributes is extracted using LLM, of which $10$ contribute to the construction of the garment structure embeddings, indicated by $*$.}
\begin{tabular}{|l|p{15cm}|}
\hline
\textbf{\textbf{Attribute}} & \textbf{Choices} \\
\hline
\textit{Category*} & T-shirt, Hoodie, Shirt, Polo, Tank, Vest, Swimsuit, Sweater, Innerwear, Windbreaker, Down Jacket, Jacket, Suit, Waistcoat, Shawl, Dress, Skirt, Knitted Coat, Leather Short Coat, Leather Long Coat, Denim Jacket, Robe, Loungewear Top, Loungewear Dress, Sports Jacket, Knitted Cardigan, Leather Jacket \\
\hline
\textit{Fit*} & Loose, Regular, Slim \\
\hline
\textit{Collar*} & Suit, Shirt, Notched, Rounded, Ruffled, Naval, Hooded, Polo, V-neck, Square, Round, Strapless, One-shoulder, Off-shoulder, Neckline, Stand-up, Baseball \\
\hline
\textit{Sleeve Length*} & Sleeveless, Short, Mid, Long, Extra Long \\
\hline
\textit{Fabric*} & Gauze, Tweed, Fur, Chiffon, Denim, PVC, Micro-Suede, Fleece, Corduroy, Knit, Lace, Synthetic, Stretch, Linen, Wool, Silk, Knitting, Leather, Velvet, Fur Blend, Coated, Mixed, Special Fabric \\
\hline
\textit{Print} & Floral, Animal, Skull, Character, Paisley, Baroque, Traditional, Cartoon, Artistic, Tech, Hand-painted, Striped, Plaid, Heart, Polka Dot, Star, Tie-dye, Camouflage, Linear, Text, Logo, Geometric, Color Block, Mixed, 3D Floral, Floral, Solid Color, Nature Scene, Objects \\
\hline
\textit{Surface Texture} & Layered, Tied, Slit, Cutout, Ruched, Pleated, Spliced, Ruffle, Contrast Stitching, Quilted, Gathered, Applique, Overlay, Hand Decorated, Beaded, Washed, Dyed, Distressed, Frayed, Printed, Splatter, Foil, Rhinestone, Flocked, Embroidered, Edge Decoration, Embossed, Punched, Knit Rib, No Craft \\
\hline
\textit{Age} & Adult, Child \\
\hline
\textit{Gender} & Female, Male \\
\hline
\textit{Length*} & Extra Short, Short, Medium, Long, Extra Long, Uncertain \\
\hline
\textit{With  Inner Wear*} & Yes, No \\
\hline
\textit{Sleeves Rolled Up*} & Yes, No \\
\hline
\textit{Top Open*} & Yes, No \\
\hline
\textit{Top Tuck In*} & Yes, No \\
\hline
\end{tabular}
\label{appendix: tab: attri}
\end{table*}

\begin{figure*}[t]
  \centering
  \includegraphics[width=1\linewidth]{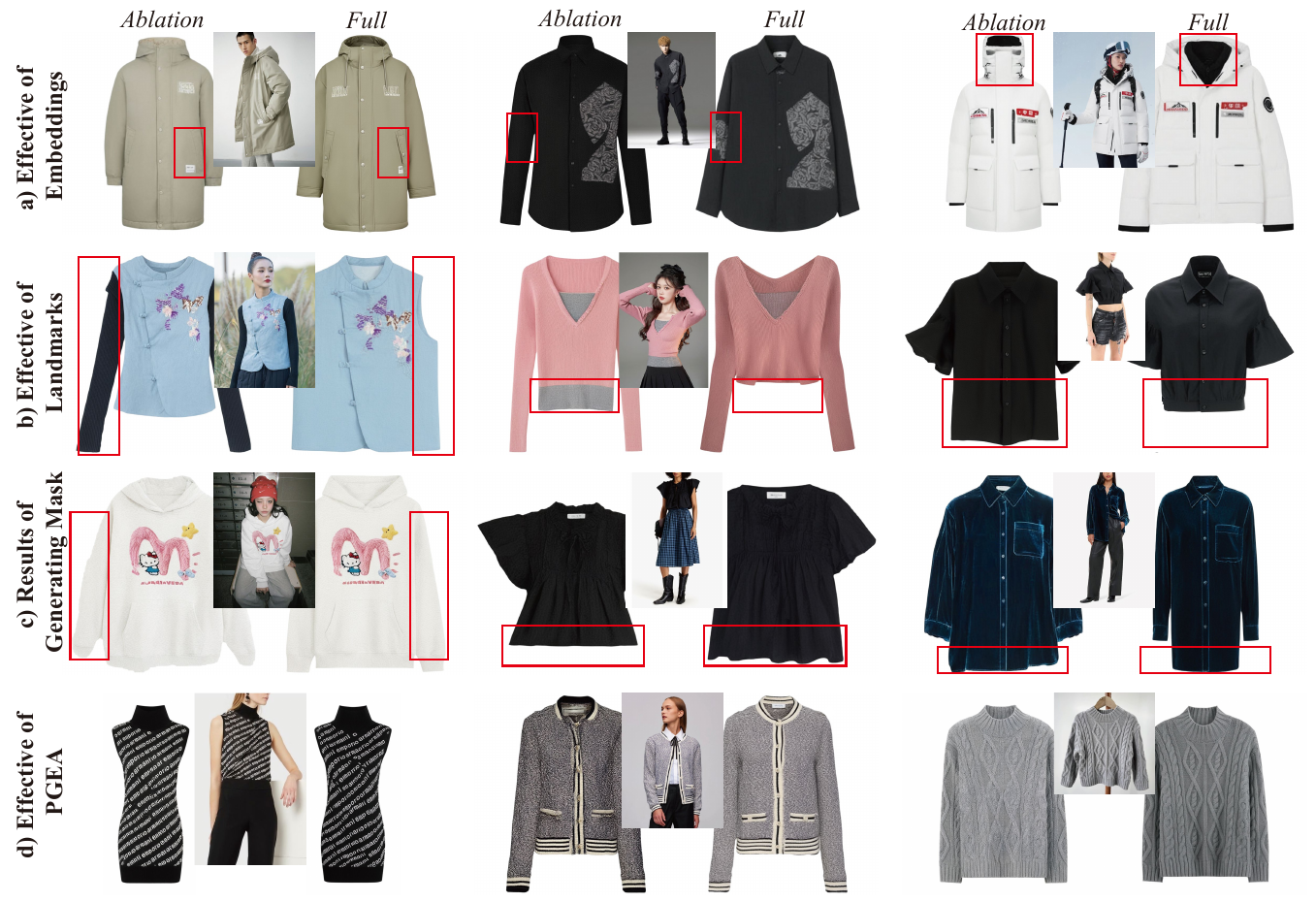}
  \caption{More visual results of ablation study.}
  \label{supp: fig: ablation_supp}
\end{figure*}

\subsection{EP-Adapter}
Inspired by IP-Adapter-plus-XL~\cite{ye2023ipadapter}, we employed a similar structure for feature information injection. The key element is that the Resampler module consists of 4 layers of MLP and 4 layers of attention mechanisms, serving as an adaptation head to modulate the structure embedding into the text embedding space. Subsequently, after encoding the values and keys, new additional cross-attention layers are integrated as follows:
\begin{equation}
\mathsf{attention}(Q, K_{\text{text}}, V_{\text{text}}) + \mathsf{attention}(Q, K_{\text{image}}, V_{\text{image}}).
\end{equation}

\paragraph{Landmark fusion.}
In the Structure LLE algorithm, the final silhouette landmark $\hat{L}_{sil}$ is fused with optimal weights $w^*$. A naive approach for integrating silhouette landmarks is to directly apply linear interpolation on the binary masks using the optimal weights $w^*$; however, this often results in blurred and oversized boundaries, which may confuse the generative model. Therefore, we adopt a compromise approach: first, we perform linear interpolation of the landmarks using the optimal weights $w^*$, and then we calculate the Intersection over Union (IOU) between the fused mask and several original landmark masks. The original mask with the highest IOU serves as the final mask to be used. This operation effectively combines information from multiple masks, yielding a result with maximum consensus while preserving clear and accurate edges that correspond to a real garment template.

\subsection{Guideline for collecting retrieval database}
To construct a memory database for retrieval, we base our selection on the training set of 65,131 standard flat-lay garments. Initially, we filter approximately 15,000 samples from the original dataset according to the principle of category balance. Subsequently, we encode the samples into embeddings and employ Density-Based Spatial Clustering of Applications with Noise (DBSCAN)~\cite{ester1996density} to cluster the samples. This step serves to eliminate potential noisy and non-standard data from the original dataset, thereby mitigating the adverse effects that outliers may have on retrieval results. We then perform random downsampling on the densely distributed samples to filter out overly similar instances. Through these operations, we aim to establish a high-quality, broadly representative standard flat-lay (embedding, garment) retrieval database with relatively low redundancy. By adjusting different clustering and downsampling parameters, we can generate retrieval libraries of four different scales: 1,000, 2,000, 4,000, and 8,000 samples, which are utilized in Section 4.2 of the main body of the paper.

\begin{figure*}[t]
  \centering
  \includegraphics[width=1\linewidth]{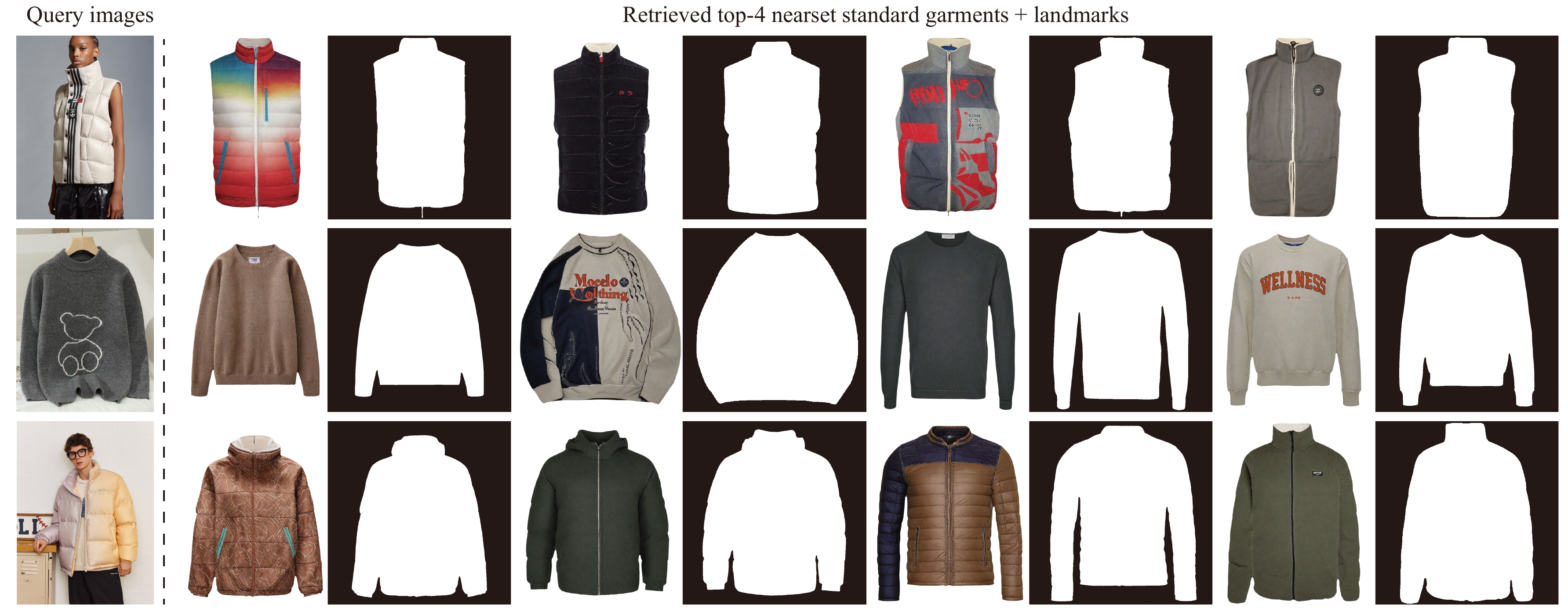}
  \caption{Retrieved results. We illustrate 4 nearest neighbor results along with their respective landmarks, given in-the-wild input. This visualization aids in intuitively understanding the effectiveness of our retrieval method.}
  \label{supp: fig: landmark_show}
\end{figure*}
 
\begin{figure}[t]
  \centering
  \includegraphics[width=1\linewidth]{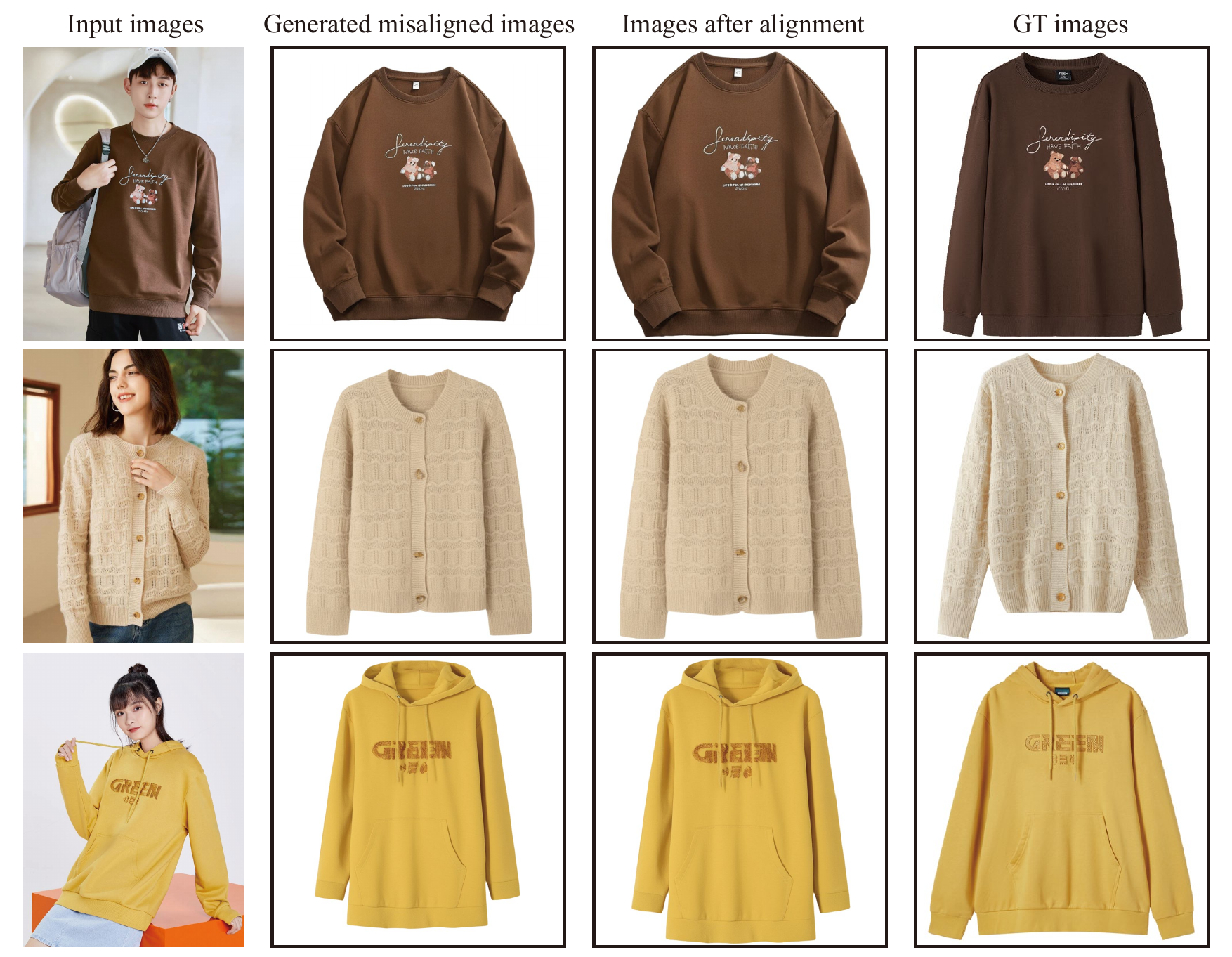}
  \caption{ We visualize the underlying reasons affecting the performance of SSIM and LPIPS including misalignment and the state of the garment. Factors such as the extent to which the sleeves are spread, the shooting angle, and lighting conditions can significantly affect the measurements of SSIM and LPIPS.}
  \label{supp: fig: ssim_show}
\end{figure}

\subsection{Dataset}
We collect a dataset named STGarment, consisting of 65,131 pairs of \emph{in-the-wild upper clothing} and \emph{standard flat lay clothing} with \emph{corresponding attributes} for training, along with 1,969 pairs for testing. The in-the-wild clothing is categorized into three main display types: clothing worn on a person, clothing laid out indoors, and clothing hung on hangers. \cref{supp: fig: datasetshow} illustrates several examples from the dataset. Before inputting the images into the network, all images are cropped or padded to ensure they are square and then resized to (768, 768).

\paragraph{Data augmentation.} 
Following~\cite{li2024anyfit}, we have implemented data augmentation techniques that could potentially enhance the model's generalization ability as well as its color accuracy performance. Specifically, the data augmentation operations include (a) horizontal flipping of images, (b) resizing standard garments and in-the-wild garments through padding (up to 10\% of the image size), (c) randomly adjusting the image's hue within a range of -5 to +5, and (d) randomly adjusting the image's contrast within a specified range (between 0.8 and 1.2 times the original contrast). Each of these operations occurs independently with a 50\% probability. Moreover, these operations are simultaneously applied to both the standard garment and in-the-wild images.

\section{Preliminary}
\label{supp: sec: preliminary}

\paragraph{Stable diffusion.}
Our RAGDiffusion is an extension of Stable Diffusion~\cite{rombach2022ldm}, which is one of the most commonly used latent diffusion models. Stable Diffusion employs a variational autoencoder~\cite{kingma2013vae} (VAE) that consists of an encoder $\mathcal{E}$ and a decoder $\mathcal{D}$ to enable image representations in the latent space. And a UNet $\epsilon_{\theta}$ is trained to denoise a Gaussian noise $\epsilon$ with a conditioning input encoded by a CLIP text encoder~\cite{radford2021clip} $\tau_{\theta}$. Given an image $\mathbf{x}$ and a text prompt $\mathbf{y}$, the training of the denoising UNet $\epsilon_{\theta}$ is performed by minimizing the following loss function:
\begin{equation}
    \mathcal{L}_{LDM} = \mathbb{E}_{\mathcal{E}(\mathbf{x}),\mathbf{y},\epsilon\sim\mathcal{N}(0, 1),t}\left[\lVert\epsilon - \epsilon_{\theta}(\mathbf{z}_t, t, \tau_{\theta}(\mathbf{y}))\rVert_2^2\right],
\end{equation}
where $t\in\{1,...,T\}$ denotes the time step of the forward diffusion process, and $\mathbf{z}_t$ is the encoded image $\mathcal{E}(\mathbf{x})$ with the added Gaussian noise $\epsilon\sim\mathcal{N}(0, 1)$ (\emph{i.e.}, the noise latent). Note that the conditioning input $\tau_{\theta}(\mathbf{y})$ is correlated with the denoising UNet by the cross-attention mechanism.

\paragraph{IP-Adapter.}
The Image Prompt Adapter (IP-Adapter) is utilized to condition the Text-to-Image (T2I) diffusion model with a reference image for style control or content indication. This is typically achieved through global control at a high-level semantic layer. Specifically, it extracts features using an image encoder (\eg, the CLIP~\cite{radford2021clip} image encoder) and incorporates an additional cross-attention layer onto the invariant text conditioning. Here, we denote \( Q \in \mathbb{R}^{N \times d} \) as the query matrices extracted from the intermediate features of the UNet, while \( K_{\text{text}} \in \mathbb{R}^{N \times d} \) and \( V_{\text{text}} \in \mathbb{R}^{N \times d} \) represent the key and value matrices derived from the prompt embeddings, where \( N \) signifies the batch size. The cross-attention layer of the text branch is computed as follows:
\begin{equation}
\mathsf{attention}(Q, K_{\text{text}}, V_{\text{text}}) = \mathsf{softmax} \left( \frac{QK_{\text{text}}^{\top}}{\sqrt{d}} \right)\cdot V_{\text{text}}.
\end{equation}
Subsequently, the IP-Adapter computes the key and value matrices \( K_{\text{image}} \in \mathbb{R}^{N \times d} \) and \( V_{\text{image}} \in \mathbb{R}^{N \times d} \) from the embedding of the reference image, and integrates the cross-attention layers as follows:
\begin{equation}
\mathsf{attention}(Q, K_{\text{text}}, V_{\text{text}}) + \mathsf{attention}(Q, K_{\text{image}}, V_{\text{image}}).
\end{equation}
During training, the weights of the original UNet are frozen, and only the projection layers of the key and value matrices in the image encoding branch, as well as the linear projection layer that maps the CLIP image embeddings, can be updated.

\begin{table}[t]
\centering
\begin{tabular}{lcccccc}
\toprule
\multicolumn{2}{c}{$K$ Num.}          && SSIM $\uparrow$ & LPIPS $\downarrow$ & FID $\downarrow$ & KID $\downarrow$ \\ 
\cmidrule{1-2} \cmidrule{4-7}  
\multicolumn{2}{l}{1}    && 0.6872 & 0.3761  & 10.27  & 1.198  \\
\multicolumn{2}{l}{2}  && 0.6915 & 0.3703  & 10.32  & 1.190  \\
\multicolumn{2}{l}{4}  && \textbf{0.6963} & \underline{0.3684}  & \textbf{9.990}  & \underline{1.092}  \\
\multicolumn{2}{l}{8}   && \underline{0.6960} & \textbf{0.3681}  & \underline{10.01}  & \textbf{1.090}  \\
\bottomrule
\end{tabular}
\caption{The number \(K\) of retrieved nearest neighbors in SLLE impacts generation performance. It's the source data for Fig. 6 in the main body of the paper.} 
\label{appendix: tab: knn_curve}
\end{table}

\section{Experiment}
\label{supp: sec: experiment}

\subsection{Alignment during evaluation protocols}
Considering that our task can essentially be viewed as reconstructing a standard flat lay from a conditioning image, we investigate the potential reasons affecting the performance of SSIM and LPIPS. As shown in~\cref{supp: fig: ssim_show}, we find that (1) the positioning of the generated flat-lay garment may influence the results of SSIM and LPIPS. Due to the lack of explicit positioning, the generated garment may deviate from the ground truth cloth in both horizontal and vertical directions, or exhibit minor differences in scale, which impacts SSIM and LPIPS. (2) Even when dealing with a standard flat-lay garment, there may still be some discrepancies in the state of the garment. Factors such as the extent to which the sleeves are spread, the shooting angle, and lighting conditions can affect the measurements of SSIM and LPIPS.

To address the measurement errors caused by misalignment (Reason 1), we systematically crop and align the generated images with the ground truth (GT) images according to the bounding boxes, as illustrated in the third column of the~\cref{supp: fig: ssim_show}. The data presented in the tables of the main body of the paper have all undergone this alignment process.

\subsection{About ControlNet baseline}
We set the in-the-wild images as the input to the ControlNet conditioning branch, due to the structural aligned mask/canny images are absent without our proposed RAG. Actually, a straightforward ControlNet cannot handle this task well, as the input in-the-wild images do not have structural alignment with the desired output images.

\subsection{Retrieval process}

We illustrate 4 nearest neighbor results along with their respective landmarks, given in-the-wild input to demonstrate the effectiveness of our retrieval method in~\cref{supp: fig: landmark_show}.  Additionally, we present precise numerical results regarding how the number \(K\) of retrieved nearest neighbors impacts generation performance in \cref{appendix: tab: knn_curve}, which serves as the source data for Fig. 6 in the main body of the paper.

\subsection{More visual results}

\cref{supp: fig: ablation_supp} provides more results about ablation study. 

\begin{figure*}[ht]
  \centering
  \includegraphics[width=0.95\linewidth]{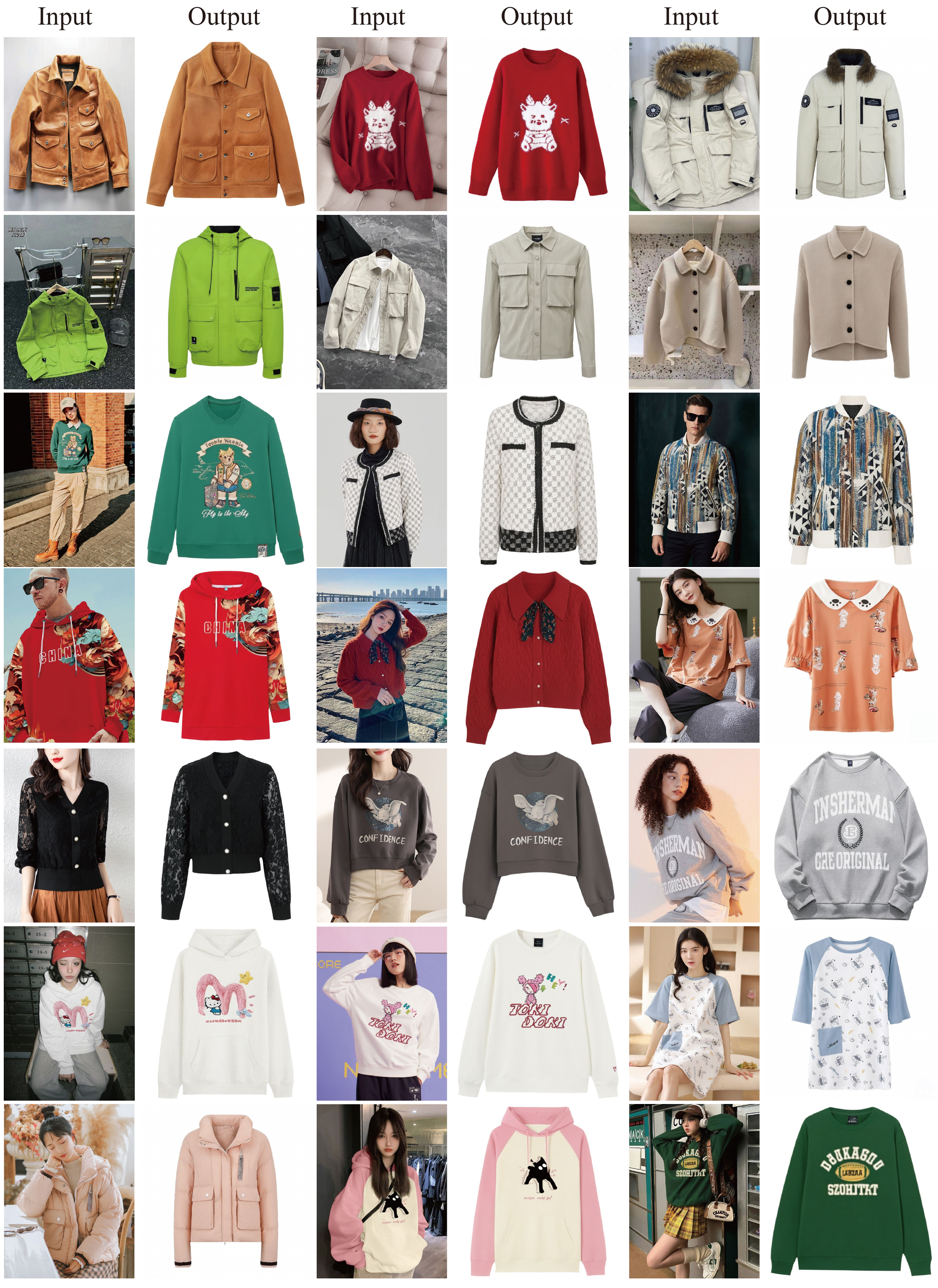}
  \vspace{-3mm}
  \caption{More visual results on the STGarment by RAGDiffusion. Best viewed when zoomed in.}
  \label{supp: fig: main}
\end{figure*}

\noindent \cref{supp: fig: main} provides more results on STGarment for inspection to demonstrate that RAGDiffusion synthesizes structurally and detail-faithful clothing assets. 

\begin{figure*}[ht]
  \centering
  \includegraphics[width=0.92\linewidth]{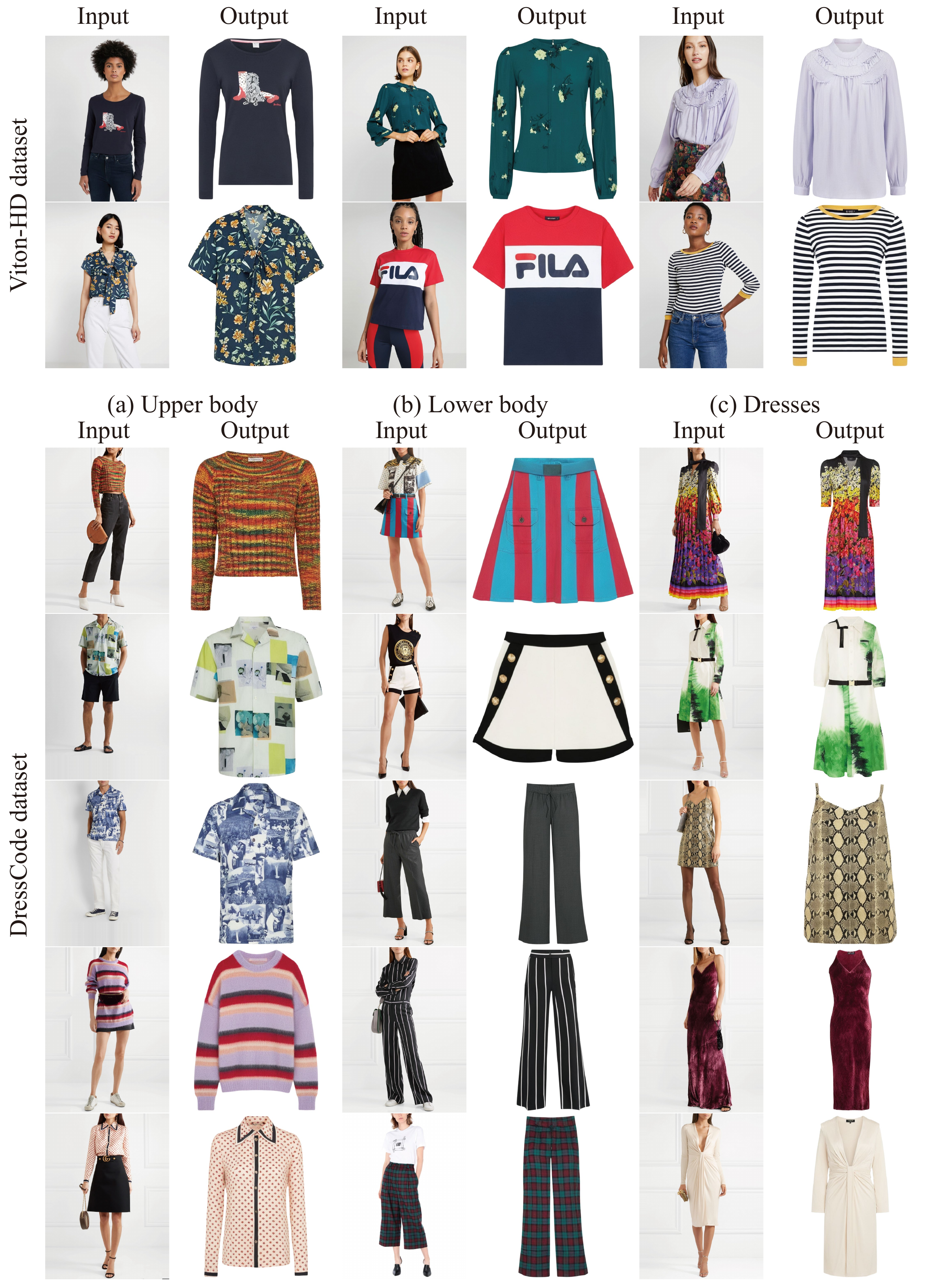}
  \vspace{-3mm}
  \caption{More cross dataset visual results on the unseen dataset Viton-HD, DressCode and the untrained categories lower-body/dresses from RAGDiffusion to validate the enhanced generalizability due to RAG. Best viewed when zoomed in.}
  \label{supp: fig: publicdata}
\end{figure*}

\noindent \cref{supp: fig: publicdata} provides more cross dataset visual results on the unseen dataset Viton-HD, DressCode and the untrained categories lower-body/dresses from RAGDiffusion to validate the enhanced generalizability due to RAG.

\clearpage

%% file: main.bbl
\begin{thebibliography}{69}
\providecommand{\natexlab}[1]{#1}
\providecommand{\url}[1]{\texttt{#1}}
\expandafter\ifx\csname urlstyle\endcsname\relax
  \providecommand{\doi}[1]{doi: #1}\else
  \providecommand{\doi}{doi: \begingroup \urlstyle{rm}\Url}\fi

\bibitem[Achiam et~al.(2023)Achiam, Adler, Agarwal, Ahmad, Akkaya, Aleman, Almeida, Altenschmidt, Altman, Anadkat, et~al.]{achiam2023gpt}
Josh Achiam, Steven Adler, Sandhini Agarwal, Lama Ahmad, Ilge Akkaya, Florencia~Leoni Aleman, Diogo Almeida, Janko Altenschmidt, Sam Altman, Shyamal Anadkat, et~al.
\newblock Gpt-4 technical report.
\newblock \emph{arXiv preprint arXiv:2303.08774}, 2023.

\bibitem[Bai et~al.(2023)Bai, Bai, Yang, Wang, Tan, Wang, Lin, Zhou, and Zhou]{bai2023qwen}
Jinze Bai, Shuai Bai, Shusheng Yang, Shijie Wang, Sinan Tan, Peng Wang, Junyang Lin, Chang Zhou, and Jingren Zhou.
\newblock Qwen-vl: A frontier large vision-language model with versatile abilities.
\newblock \emph{arXiv preprint arXiv:2308.12966}, 2023.

\bibitem[Blanz and Vetter(2003)]{blanz2003face}
Volker Blanz and Thomas Vetter.
\newblock Face recognition based on fitting a 3d morphable model.
\newblock \emph{IEEE Transactions on pattern analysis and machine intelligence}, 25\penalty0 (9):\penalty0 1063--1074, 2003.

\bibitem[Blattmann et~al.(2022)Blattmann, Rombach, Oktay, M{\"u}ller, and Ommer]{blattmann2022retrievaldiff}
Andreas Blattmann, Robin Rombach, Kaan Oktay, Jonas M{\"u}ller, and Bj{\"o}rn Ommer.
\newblock Retrieval-augmented diffusion models.
\newblock \emph{Advances in Neural Information Processing Systems}, 35:\penalty0 15309--15324, 2022.

\bibitem[Borgeaud et~al.(2022)Borgeaud, Mensch, Hoffmann, Cai, Rutherford, Millican, Van Den~Driessche, Lespiau, Damoc, Clark, et~al.]{RETRO}
Sebastian Borgeaud, Arthur Mensch, Jordan Hoffmann, Trevor Cai, Eliza Rutherford, Katie Millican, George~Bm Van Den~Driessche, Jean-Baptiste Lespiau, Bogdan Damoc, Aidan Clark, et~al.
\newblock Improving language models by retrieving from trillions of tokens.
\newblock In \emph{International conference on machine learning}, pages 2206--2240. PMLR, 2022.

\bibitem[Borji(2023)]{borji2023qualitative}
Ali Borji.
\newblock Qualitative failures of image generation models and their application in detecting deepfakes.
\newblock \emph{Image and Vision Computing}, 137:\penalty0 104771, 2023.

\bibitem[Casanova et~al.(2021)Casanova, Careil, Verbeek, Drozdzal, and Romero~Soriano]{casanova2021instance}
Arantxa Casanova, Marlene Careil, Jakob Verbeek, Michal Drozdzal, and Adriana Romero~Soriano.
\newblock Instance-conditioned gan.
\newblock \emph{Advances in Neural Information Processing Systems}, 34:\penalty0 27517--27529, 2021.

\bibitem[Chang et~al.(2023)Chang, Shi, Gao, Fu, Xu, Song, Yan, Yang, and Soleymani]{chang2023magicdance}
Di Chang, Yichun Shi, Quankai Gao, Jessica Fu, Hongyi Xu, Guoxian Song, Qing Yan, Xiao Yang, and Mohammad Soleymani.
\newblock Magicdance: Realistic human dance video generation with motions \& facial expressions transfer.
\newblock \emph{arXiv preprint arXiv:2311.12052}, 2023.

\bibitem[Chen et~al.(2024{\natexlab{a}})Chen, Laina, and Vedaldi]{chen2024layoutcontrol}
Minghao Chen, Iro Laina, and Andrea Vedaldi.
\newblock Training-free layout control with cross-attention guidance.
\newblock In \emph{Proceedings of the IEEE/CVF Winter Conference on Applications of Computer Vision}, pages 5343--5353, 2024{\natexlab{a}}.

\bibitem[Chen et~al.(2020)Chen, Kornblith, Norouzi, and Hinton]{chen2020simclr}
Ting Chen, Simon Kornblith, Mohammad Norouzi, and Geoffrey Hinton.
\newblock A simple framework for contrastive learning of visual representations.
\newblock In \emph{International conference on machine learning}, pages 1597--1607. PMLR, 2020.

\bibitem[Chen et~al.(2024{\natexlab{b}})Chen, Feng, Chen, Wang, Zhang, Liu, Shen, and Zhao]{chen2024zero}
Xi Chen, Yutong Feng, Mengting Chen, Yiyang Wang, Shilong Zhang, Yu Liu, Yujun Shen, and Hengshuang Zhao.
\newblock Zero-shot image editing with reference imitation.
\newblock \emph{arXiv preprint arXiv:2406.07547}, 2024{\natexlab{b}}.

\bibitem[Choi et~al.(2021)Choi, Park, Lee, and Choo]{choi2021vitonhd}
Seunghwan Choi, Sunghyun Park, Minsoo Lee, and Jaegul Choo.
\newblock Viton-hd: High-resolution virtual try-on via misalignment-aware normalization.
\newblock In \emph{Proceedings of the IEEE/CVF conference on computer vision and pattern recognition}, 2021.

\bibitem[Ding et~al.(2020)Ding, Ma, Wang, and Simoncelli]{dists}
Keyan Ding, Kede Ma, Shiqi Wang, and Eero~P Simoncelli.
\newblock Image quality assessment: Unifying structure and texture similarity.
\newblock \emph{IEEE transactions on pattern analysis and machine intelligence}, 44\penalty0 (5):\penalty0 2567--2581, 2020.

\bibitem[Dosovitskiy(2020)]{dosovitskiy2020vit}
Alexey Dosovitskiy.
\newblock An image is worth 16x16 words: Transformers for image recognition at scale.
\newblock \emph{arXiv preprint arXiv:2010.11929}, 2020.

\bibitem[Douze et~al.(2024)Douze, Guzhva, Deng, Johnson, Szilvasy, Mazaré, Lomeli, Hosseini, and Jégou]{douze2024faiss}
Matthijs Douze, Alexandr Guzhva, Chengqi Deng, Jeff Johnson, Gergely Szilvasy, Pierre-Emmanuel Mazaré, Maria Lomeli, Lucas Hosseini, and Hervé Jégou.
\newblock The faiss library.
\newblock 2024.

\bibitem[Ester et~al.(1996)Ester, Kriegel, Sander, Xu, et~al.]{ester1996density}
Martin Ester, Hans-Peter Kriegel, J{\"o}rg Sander, Xiaowei Xu, et~al.
\newblock A density-based algorithm for discovering clusters in large spatial databases with noise.
\newblock In \emph{kdd}, pages 226--231, 1996.

\bibitem[forest labs(2024)]{flux}
Black forest labs.
\newblock Flux.1-dev.
\newblock \url{https://github.com/black-forest-labs/flux}, 2024.

\bibitem[Goodfellow et~al.(2014)Goodfellow, Pouget-Abadie, Mirza, Xu, Warde-Farley, Ozair, Courville, and Bengio]{goodfellow2014GAN}
Ian Goodfellow, Jean Pouget-Abadie, Mehdi Mirza, Bing Xu, David Warde-Farley, Sherjil Ozair, Aaron Courville, and Yoshua Bengio.
\newblock Generative adversarial nets.
\newblock \emph{Advances in neural information processing systems}, 2014.

\bibitem[Guo et~al.(2003)Guo, Wang, Bell, Bi, and Greer]{guo2003knn}
Gongde Guo, Hui Wang, David Bell, Yaxin Bi, and Kieran Greer.
\newblock Knn model-based approach in classification.
\newblock In \emph{On The Move to Meaningful Internet Systems 2003: CoopIS, DOA, and ODBASE: OTM Confederated International Conferences, CoopIS, DOA, and ODBASE 2003, Catania, Sicily, Italy, November 3-7, 2003. Proceedings}, pages 986--996. Springer, 2003.

\bibitem[He et~al.(2020)He, Fan, Wu, Xie, and Girshick]{he2020moco}
Kaiming He, Haoqi Fan, Yuxin Wu, Saining Xie, and Ross Girshick.
\newblock Momentum contrast for unsupervised visual representation learning.
\newblock In \emph{Proceedings of the IEEE/CVF conference on computer vision and pattern recognition}, pages 9729--9738, 2020.

\bibitem[He et~al.(2024)He, Yao, Zhang, Yu, Liu, and Xu]{he2024dresscode}
Kai He, Kaixin Yao, Qixuan Zhang, Jingyi Yu, Lingjie Liu, and Lan Xu.
\newblock Dresscode: Autoregressively sewing and generating garments from text guidance.
\newblock \emph{ACM Transactions on Graphics (TOG)}, 43\penalty0 (4):\penalty0 1--13, 2024.

\bibitem[Ho et~al.(2020)Ho, Jain, and Abbeel]{ho2020ddpm}
Jonathan Ho, Ajay Jain, and Pieter Abbeel.
\newblock Denoising diffusion probabilistic models.
\newblock \emph{Advances in neural information processing systems}, 2020.

\bibitem[Hu et~al.(2023{\natexlab{a}})Hu, Gao, Zhang, Sun, Zhang, and Bo]{hu2023animate}
Li Hu, Xin Gao, Peng Zhang, Ke Sun, Bang Zhang, and Liefeng Bo.
\newblock Animate anyone: Consistent and controllable image-to-video synthesis for character animation.
\newblock \emph{arXiv preprint arXiv:2311.17117}, 2023{\natexlab{a}}.

\bibitem[Hu et~al.(2023{\natexlab{b}})Hu, Zheng, Liu, Zheng, Wang, Tao, and Cham]{hu2023cocktail}
Minghui Hu, Jianbin Zheng, Daqing Liu, Chuanxia Zheng, Chaoyue Wang, Dacheng Tao, and Tat-Jen Cham.
\newblock Cocktail: Mixing multi-modality control for text-conditional image generation.
\newblock In \emph{Thirty-seventh Conference on Neural Information Processing Systems}, 2023{\natexlab{b}}.

\bibitem[Ju et~al.(2023)Ju, Zeng, Zhao, Wang, Zhang, and Xu]{ju2023humansd}
Xuan Ju, Ailing Zeng, Chenchen Zhao, Jianan Wang, Lei Zhang, and Qiang Xu.
\newblock Humansd: A native skeleton-guided diffusion model for human image generation.
\newblock In \emph{Proceedings of the IEEE/CVF International Conference on Computer Vision}, pages 15988--15998, 2023.

\bibitem[Karras et~al.(2019)Karras, Laine, and Aila]{karras2019style}
Tero Karras, Samuli Laine, and Timo Aila.
\newblock A style-based generator architecture for generative adversarial networks.
\newblock In \emph{Proceedings of the IEEE/CVF conference on computer vision and pattern recognition}, pages 4401--4410, 2019.

\bibitem[Kingma and Welling(2013)]{kingma2013vae}
Diederik~P Kingma and Max Welling.
\newblock Auto-encoding variational bayes.
\newblock \emph{arXiv preprint arXiv:1312.6114}, 2013.

\bibitem[Li et~al.(2023)Li, Liu, Wu, Mu, Yang, Gao, Li, and Lee]{li2023gligen}
Yuheng Li, Haotian Liu, Qingyang Wu, Fangzhou Mu, Jianwei Yang, Jianfeng Gao, Chunyuan Li, and Yong~Jae Lee.
\newblock Gligen: Open-set grounded text-to-image generation.
\newblock In \emph{Proceedings of the IEEE/CVF Conference on Computer Vision and Pattern Recognition}, pages 22511--22521, 2023.

\bibitem[Li et~al.(2024)Li, Zhou, Shang, Lin, Chen, and Ni]{li2024anyfit}
Yuhan Li, Hao Zhou, Wenxiang Shang, Ran Lin, Xuanhong Chen, and Bingbing Ni.
\newblock Anyfit: Controllable virtual try-on for any combination of attire across any scenario.
\newblock \emph{arXiv preprint arXiv:2405.18172}, 2024.

\bibitem[Liu et~al.(2023)Liu, Kortylewski, Bai, Bai, and Yuille]{liu2023intriguing}
Qihao Liu, Adam Kortylewski, Yutong Bai, Song Bai, and Alan Yuille.
\newblock Intriguing properties of text-guided diffusion models.
\newblock \emph{arXiv preprint arXiv:2306.00974}, 2, 2023.

\bibitem[Microsoft(2023)]{deepspeed}
Microsoft.
\newblock Deepspeed.
\newblock \url{https://github.com/microsoft/DeepSpeed}, 2023.

\bibitem[Morelli et~al.(2023)Morelli, Baldrati, Cartella, Cornia, Bertini, and Cucchiara]{morelli2023ladi}
Davide Morelli, Alberto Baldrati, Giuseppe Cartella, Marcella Cornia, Marco Bertini, and Rita Cucchiara.
\newblock {LaDI-VTON: Latent Diffusion Textual-Inversion Enhanced Virtual Try-On}.
\newblock In \emph{Proceedings of the ACM International Conference on Multimedia}, 2023.

\bibitem[Mou et~al.(2024)Mou, Wang, Xie, Wu, Zhang, Qi, and Shan]{mou2024t2iadapter}
Chong Mou, Xintao Wang, Liangbin Xie, Yanze Wu, Jian Zhang, Zhongang Qi, and Ying Shan.
\newblock T2i-adapter: Learning adapters to dig out more controllable ability for text-to-image diffusion models.
\newblock In \emph{Proceedings of the AAAI Conference on Artificial Intelligence}, pages 4296--4304, 2024.

\bibitem[Oquab et~al.(2023)Oquab, Darcet, Moutakanni, Vo, Szafraniec, Khalidov, Fernandez, Haziza, Massa, El-Nouby, Howes, Huang, Xu, Sharma, Li, Galuba, Rabbat, Assran, Ballas, Synnaeve, Misra, Jegou, Mairal, Labatut, Joulin, and Bojanowski]{oquab2023dinov2}
Maxime Oquab, Timothée Darcet, Theo Moutakanni, Huy~V. Vo, Marc Szafraniec, Vasil Khalidov, Pierre Fernandez, Daniel Haziza, Francisco Massa, Alaaeldin El-Nouby, Russell Howes, Po-Yao Huang, Hu Xu, Vasu Sharma, Shang-Wen Li, Wojciech Galuba, Mike Rabbat, Mido Assran, Nicolas Ballas, Gabriel Synnaeve, Ishan Misra, Herve Jegou, Julien Mairal, Patrick Labatut, Armand Joulin, and Piotr Bojanowski.
\newblock Dinov2: Learning robust visual features without supervision, 2023.

\bibitem[Parmar et~al.(2022)Parmar, Zhang, and Zhu]{parmar2021cleanfid}
Gaurav Parmar, Richard Zhang, and Jun-Yan Zhu.
\newblock On aliased resizing and surprising subtleties in gan evaluation.
\newblock In \emph{Proceedings of the IEEE/CVF Conference on Computer Vision and Pattern Recognition}, 2022.

\bibitem[Peebles and Xie(2023)]{peebles2023dit}
William Peebles and Saining Xie.
\newblock Scalable diffusion models with transformers.
\newblock In \emph{Proceedings of the IEEE/CVF International Conference on Computer Vision}, pages 4195--4205, 2023.

\bibitem[Podell et~al.(2023)Podell, English, Lacey, Blattmann, Dockhorn, M{\"u}ller, Penna, and Rombach]{podell2023sdxl}
Dustin Podell, Zion English, Kyle Lacey, Andreas Blattmann, Tim Dockhorn, Jonas M{\"u}ller, Joe Penna, and Robin Rombach.
\newblock Sdxl: Improving latent diffusion models for high-resolution image synthesis.
\newblock \emph{arXiv preprint arXiv:2307.01952}, 2023.

\bibitem[Qin et~al.(2023)Qin, Zhang, Yu, Feng, Yang, Zhou, Wang, Niebles, Xiong, Savarese, et~al.]{qin2023unicontrol}
Can Qin, Shu Zhang, Ning Yu, Yihao Feng, Xinyi Yang, Yingbo Zhou, Huan Wang, Juan~Carlos Niebles, Caiming Xiong, Silvio Savarese, et~al.
\newblock Unicontrol: A unified diffusion model for controllable visual generation in the wild.
\newblock \emph{arXiv preprint arXiv:2305.11147}, 2023.

\bibitem[Radford et~al.(2021)Radford, Kim, Hallacy, Ramesh, Goh, Agarwal, Sastry, Askell, Mishkin, Clark, et~al.]{radford2021clip}
Alec Radford, Jong~Wook Kim, Chris Hallacy, Aditya Ramesh, Gabriel Goh, Sandhini Agarwal, Girish Sastry, Amanda Askell, Pamela Mishkin, Jack Clark, et~al.
\newblock Learning transferable visual models from natural language supervision.
\newblock In \emph{International conference on machine learning}, 2021.

\bibitem[Ram et~al.(2023)Ram, Levine, Dalmedigos, Muhlgay, Shashua, Leyton-Brown, and Shoham]{ram2023conrag2}
Ori Ram, Yoav Levine, Itay Dalmedigos, Dor Muhlgay, Amnon Shashua, Kevin Leyton-Brown, and Yoav Shoham.
\newblock In-context retrieval-augmented language models.
\newblock \emph{Transactions of the Association for Computational Linguistics}, 11:\penalty0 1316--1331, 2023.

\bibitem[Rombach et~al.(2022{\natexlab{a}})Rombach, Blattmann, Lorenz, Esser, and Ommer]{rombach2022ldm}
Robin Rombach, Andreas Blattmann, Dominik Lorenz, Patrick Esser, and Bj{\"o}rn Ommer.
\newblock High-resolution image synthesis with latent diffusion models.
\newblock In \emph{Proceedings of the IEEE/CVF conference on computer vision and pattern recognition}, 2022{\natexlab{a}}.

\bibitem[Rombach et~al.(2022{\natexlab{b}})Rombach, Blattmann, and Ommer]{rombach2022retrievalArtistic}
Robin Rombach, Andreas Blattmann, and Bj{\"o}rn Ommer.
\newblock Text-guided synthesis of artistic images with retrieval-augmented diffusion models.
\newblock \emph{arXiv preprint arXiv:2207.13038}, 2022{\natexlab{b}}.

\bibitem[Roweis and Saul(2000)]{roweis2000lle}
Sam~T Roweis and Lawrence~K Saul.
\newblock Nonlinear dimensionality reduction by locally linear embedding.
\newblock \emph{science}, 290\penalty0 (5500):\penalty0 2323--2326, 2000.

\bibitem[Seo et~al.(2024)Seo, Hong, Jang, Kim, Kwak, Lee, and Kim]{seo2024retrieval3d}
Junyoung Seo, Susung Hong, Wooseok Jang, In{\`e}s~Hyeonsu Kim, Minseop Kwak, Doyup Lee, and Seungryong Kim.
\newblock Retrieval-augmented score distillation for text-to-3d generation.
\newblock \emph{arXiv preprint arXiv:2402.02972}, 2024.

\bibitem[Shen et~al.(2024)Shen, Huang, and Wang]{shen2024igr}
Le Shen, Rong Huang, and Zhijie Wang.
\newblock Igr: Improving diffusion model for garment restoration from person image.
\newblock \emph{arXiv preprint arXiv:2412.11513}, 2024.

\bibitem[Sheynin et~al.(2022)Sheynin, Ashual, Polyak, Singer, Gafni, Nachmani, and Taigman]{sheynin2022knndiffusion}
Shelly Sheynin, Oron Ashual, Adam Polyak, Uriel Singer, Oran Gafni, Eliya Nachmani, and Yaniv Taigman.
\newblock Knn-diffusion: Image generation via large-scale retrieval.
\newblock \emph{arXiv preprint arXiv:2204.02849}, 2022.

\bibitem[Song et~al.(2020)Song, Meng, and Ermon]{song2020ddim}
Jiaming Song, Chenlin Meng, and Stefano Ermon.
\newblock Denoising diffusion implicit models.
\newblock \emph{arXiv preprint arXiv:2010.02502}, 2020.

\bibitem[Sutherland et~al.(2018)Sutherland, Arbel, and Gretton]{sutherland2018kid}
JD Sutherland, Michael Arbel, and Arthur Gretton.
\newblock Demystifying mmd gans.
\newblock In \emph{International Conference for Learning Representations}, 2018.

\bibitem[Tseng et~al.(2020)Tseng, Lee, Jiang, Yang, and Yang]{tseng2020retrievegan}
Hung-Yu Tseng, Hsin-Ying Lee, Lu Jiang, Ming-Hsuan Yang, and Weilong Yang.
\newblock Retrievegan: Image synthesis via differentiable patch retrieval.
\newblock In \emph{Computer Vision--ECCV 2020: 16th European Conference, Glasgow, UK, August 23--28, 2020, Proceedings, Part VIII 16}, pages 242--257. Springer, 2020.

\bibitem[Van~der Maaten and Hinton(2008)]{van2008visualizing}
Laurens Van~der Maaten and Geoffrey Hinton.
\newblock Visualizing data using t-sne.
\newblock \emph{Journal of machine learning research}, 9\penalty0 (11), 2008.

\bibitem[Velioglu et~al.(2024)Velioglu, Bevandic, Chan, and Hammer]{velioglu2024tryoffdiff}
Riza Velioglu, Petra Bevandic, Robin Chan, and Barbara Hammer.
\newblock Tryoffdiff: Virtual-try-off via high-fidelity garment reconstruction using diffusion models.
\newblock \emph{arXiv preprint arXiv:2411.18350}, 2024.

\bibitem[Wang et~al.(2004)Wang, Bovik, Sheikh, and Simoncelli]{wang2004ssim}
Zhou Wang, Alan~C Bovik, Hamid~R Sheikh, and Eero~P Simoncelli.
\newblock Image quality assessment: from error visibility to structural similarity.
\newblock \emph{IEEE transactions on image processing}, 2004.

\bibitem[Xarchakos and Koukopoulos(2024)]{xarchakos2024tryoffanyone}
Ioannis Xarchakos and Theodoros Koukopoulos.
\newblock Tryoffanyone: Tiled cloth generation from a dressed person.
\newblock \emph{arXiv preprint arXiv:2412.08573}, 2024.

\bibitem[Xie et~al.(2023)Xie, Li, Huang, Liu, Zhang, Zheng, and Shou]{xie2023boxdiff}
Jinheng Xie, Yuexiang Li, Yawen Huang, Haozhe Liu, Wentian Zhang, Yefeng Zheng, and Mike~Zheng Shou.
\newblock Boxdiff: Text-to-image synthesis with training-free box-constrained diffusion.
\newblock In \emph{Proceedings of the IEEE/CVF International Conference on Computer Vision}, pages 7452--7461, 2023.

\bibitem[Yang et~al.(2023{\natexlab{a}})Yang, Gu, Zhang, Zhang, Chen, Sun, Chen, and Wen]{yang2023paintbyexample}
Binxin Yang, Shuyang Gu, Bo Zhang, Ting Zhang, Xuejin Chen, Xiaoyan Sun, Dong Chen, and Fang Wen.
\newblock Paint by example: Exemplar-based image editing with diffusion models.
\newblock In \emph{Proceedings of the IEEE/CVF Conference on Computer Vision and Pattern Recognition}, 2023{\natexlab{a}}.

\bibitem[Yang et~al.(2023{\natexlab{b}})Yang, Wang, Gan, Li, Lin, Wu, Duan, Liu, Liu, Zeng, et~al.]{yang2023reco}
Zhengyuan Yang, Jianfeng Wang, Zhe Gan, Linjie Li, Kevin Lin, Chenfei Wu, Nan Duan, Zicheng Liu, Ce Liu, Michael Zeng, et~al.
\newblock Reco: Region-controlled text-to-image generation.
\newblock In \emph{Proceedings of the IEEE/CVF Conference on Computer Vision and Pattern Recognition}, pages 14246--14255, 2023{\natexlab{b}}.

\bibitem[Ye et~al.(2023{\natexlab{a}})Ye, Zhang, Liu, Han, and Yang]{ye2023ipadapter}
Hu Ye, Jun Zhang, Sibo Liu, Xiao Han, and Wei Yang.
\newblock Ip-adapter: Text compatible image prompt adapter for text-to-image diffusion models.
\newblock \emph{arXiv preprint arXiv:2308.06721}, 2023{\natexlab{a}}.

\bibitem[Ye et~al.(2023{\natexlab{b}})Ye, He, Jiang, Huang, Huang, Liu, Ren, Yin, Ma, and Zhao]{ye2023geneface++}
Zhenhui Ye, Jinzheng He, Ziyue Jiang, Rongjie Huang, Jiawei Huang, Jinglin Liu, Yi Ren, Xiang Yin, Zejun Ma, and Zhou Zhao.
\newblock Geneface++: Generalized and stable real-time audio-driven 3d talking face generation.
\newblock \emph{arXiv preprint arXiv:2305.00787}, 2023{\natexlab{b}}.

\bibitem[Zeng et~al.(2020)Zeng, Zhao, Gao, and Zhang]{zeng2020tilegan}
Wei Zeng, Mingbo Zhao, Yuan Gao, and Zhao Zhang.
\newblock Tilegan: category-oriented attention-based high-quality tiled clothes generation from dressed person.
\newblock \emph{Neural Computing and Applications}, 32:\penalty0 17587--17600, 2020.

\bibitem[Zhai et~al.(2023)Zhai, Mustafa, Kolesnikov, and Beyer]{zhai2023sigmoid}
Xiaohua Zhai, Basil Mustafa, Alexander Kolesnikov, and Lucas Beyer.
\newblock Sigmoid loss for language image pre-training.
\newblock In \emph{Proceedings of the IEEE/CVF international conference on computer vision}, pages 11975--11986, 2023.

\bibitem[Zhang et~al.(2023{\natexlab{a}})Zhang, Rao, and Agrawala]{zhang2023controlnet}
Lvmin Zhang, Anyi Rao, and Maneesh Agrawala.
\newblock Adding conditional control to text-to-image diffusion models.
\newblock In \emph{Proceedings of the IEEE/CVF International Conference on Computer Vision}, pages 3836--3847, 2023{\natexlab{a}}.

\bibitem[Zhang et~al.(2023{\natexlab{b}})Zhang, Guo, Pan, Cai, Hong, Li, Yang, and Liu]{zhang2023remodiffuse}
Mingyuan Zhang, Xinying Guo, Liang Pan, Zhongang Cai, Fangzhou Hong, Huirong Li, Lei Yang, and Ziwei Liu.
\newblock Remodiffuse: Retrieval-augmented motion diffusion model.
\newblock In \emph{Proceedings of the IEEE/CVF International Conference on Computer Vision}, pages 364--373, 2023{\natexlab{b}}.

\bibitem[Zhang et~al.(2018)Zhang, Isola, Efros, Shechtman, and Wang]{zhang2018perceptual}
Richard Zhang, Phillip Isola, Alexei~A Efros, Eli Shechtman, and Oliver Wang.
\newblock The unreasonable effectiveness of deep features as a perceptual metric.
\newblock In \emph{Proceedings of the IEEE conference on computer vision and pattern recognition}, 2018.

\bibitem[Zhang et~al.(2024{\natexlab{a}})Zhang, Chong, Zhang, Li, Cheng, Yan, and Liang]{zhang2024garmentaligner}
Shiyue Zhang, Zheng Chong, Xujie Zhang, Hanhui Li, Yuhao Cheng, Yiqiang Yan, and Xiaodan Liang.
\newblock Garmentaligner: Text-to-garment generation via retrieval-augmented multi-level corrections.
\newblock \emph{arXiv preprint arXiv:2408.12352}, 2024{\natexlab{a}}.

\bibitem[Zhang et~al.(2023{\natexlab{c}})Zhang, Zhang, Vineet, Joshi, and Wang]{zhang2023controllablegpt4}
Tianjun Zhang, Yi Zhang, Vibhav Vineet, Neel Joshi, and Xin Wang.
\newblock Controllable text-to-image generation with gpt-4.
\newblock \emph{arXiv preprint arXiv:2305.18583}, 2023{\natexlab{c}}.

\bibitem[Zhang et~al.(2022)Zhang, Sha, Kampffmeyer, Xie, Jie, Huang, Peng, and Liang]{zhang2022armani}
Xujie Zhang, Yu Sha, Michael~C Kampffmeyer, Zhenyu Xie, Zequn Jie, Chengwen Huang, Jianqing Peng, and Xiaodan Liang.
\newblock Armani: Part-level garment-text alignment for unified cross-modal fashion design.
\newblock In \emph{Proceedings of the 30th ACM International Conference on Multimedia}, pages 4525--4535, 2022.

\bibitem[Zhang et~al.(2023{\natexlab{d}})Zhang, Yang, Kampffmeyer, Zhang, Zhang, Lu, Lin, Xu, and Liang]{zhang2023diffcloth}
Xujie Zhang, Binbin Yang, Michael~C Kampffmeyer, Wenqing Zhang, Shiyue Zhang, Guansong Lu, Liang Lin, Hang Xu, and Xiaodan Liang.
\newblock Diffcloth: Diffusion based garment synthesis and manipulation via structural cross-modal semantic alignment.
\newblock In \emph{Proceedings of the IEEE/CVF International Conference on Computer Vision}, pages 23154--23163, 2023{\natexlab{d}}.

\bibitem[Zhang et~al.(2024{\natexlab{b}})Zhang, Song, Zhan, Chen, Xu, Luo, Zhang, and Liu]{zhang2024boow}
Xuanpu Zhang, Dan Song, Pengxin Zhan, Qingguo Chen, Zhao Xu, Weihua Luo, Kaifu Zhang, and Anan Liu.
\newblock Boow-vton: Boosting in-the-wild virtual try-on via mask-free pseudo data training.
\newblock \emph{arXiv preprint arXiv:2408.06047}, 2024{\natexlab{b}}.

\bibitem[Zhao et~al.(2024)Zhao, Chen, Chen, Bao, Hao, Yuan, and Wong]{zhao2024uni}
Shihao Zhao, Dongdong Chen, Yen-Chun Chen, Jianmin Bao, Shaozhe Hao, Lu Yuan, and Kwan-Yee~K Wong.
\newblock Uni-controlnet: All-in-one control to text-to-image diffusion models.
\newblock \emph{Advances in Neural Information Processing Systems}, 36, 2024.

\end{thebibliography}
